\pdfoutput=1
\documentclass[11pt]{article}

\usepackage[final]{acl}

\usepackage{times}
\usepackage{latexsym}
\usepackage{multirow}
\usepackage{float}
\usepackage{graphicx}
\usepackage{longtable}
\usepackage{array}
\usepackage{makecell}
\usepackage{colortbl}
\usepackage{xcolor}
\usepackage{booktabs}
\definecolor{darkred}{RGB}{139, 0, 0}
\definecolor{darkblue}{RGB}{0,55,98}
\usepackage{mathtools}
\usepackage{amsmath,amsfonts,bm}

\def\eqref#1{equation~\ref{#1}}
\def\1{\bm{1}}

\def\ry{{\textnormal{y}}}

\DeclareMathAlphabet{\mathsfit}{\encodingdefault}{\sfdefault}{m}{sl}
\SetMathAlphabet{\mathsfit}{bold}{\encodingdefault}{\sfdefault}{bx}{n}

\def\gC{{\mathcal{C}}}
\def\gD{{\mathcal{D}}}

\def\gI{{\mathcal{I}}}
\def\gJ{{\mathcal{J}}}

\def\gM{{\mathcal{M}}}

\def\gY{{\mathcal{Y}}}

\usepackage[T1]{fontenc}
\usepackage[utf8]{inputenc}

\usepackage{microtype}

\usepackage{inconsolata}

\usepackage{graphicx}
\graphicspath{figs/}

\title{MineAgent: Towards Remote-Sensing Mineral Exploration with Multimodal Large Language Models}

\author{
 \textbf{Beibei Yu\textsuperscript{1}},
 \textbf{Tao Shen\textsuperscript{1}},
 \textbf{Hongbin Na\textsuperscript{1}},
 \textbf{Ling Chen\textsuperscript{1}},
 \textbf{Denqi Li\textsuperscript{2}},
\\
 \textsuperscript{1}Australian Artificial Intelligence Institute, University of Technology Sydney,
 \\
 \textsuperscript{2}Faculty of Science and Engineering,  Curtin University
}

\begin{document}
\maketitle
\begin{abstract}
Remote-sensing mineral exploration is critical for identifying economically viable mineral deposits, yet it poses significant challenges for multimodal large language models (MLLMs). These include limitations in domain-specific geological knowledge and difficulties in reasoning across multiple remote-sensing images, further exacerbating long-context issues. To address these, we present MineAgent, a modular framework leveraging hierarchical judging and decision-making modules to improve multi-image reasoning and spatial-spectral integration. Complementing this, we propose MineBench, a benchmark specific for evaluating MLLMs in domain-specific mineral exploration tasks using geological and hyperspectral data. Extensive experiments demonstrate the effectiveness of MineAgent, highlighting its potential to advance MLLMs in remote-sensing mineral exploration. 
% Mineral exploration is critical for discovering valuable mineral deposits. While geologists increasingly employ deep learning on remote sensing data to enhance exploration efficiency, the potential of Multi-modal Large Language Models (MLLMs) in mineral exploration remains largely unexplored. We introduce MineAgent, a novel MLLM agent framework designed to automate mineral exploration without requiring training on domain-specific data.
% To tackle the complexity of mineral exploration, we divide the exploration pipeline into three specialized modules under a local-global visual reasoning framework. Additionally, a collaborative evaluation module integrates intermediate results to minimize cascading errors. To overcome benchmark limitations, we introduce MineBench, a dataset of 612 human-verified geological areas for evaluating MineAgent. Experimental results show that MineAgent improves performance across various MLLMs. However, performance declines significantly with increasing task complexity, underscoring challenges in applying MLLMs to complex geospatial tasks.

\end{abstract}

\section{Introduction}

Mineral exploration is a systematic geological investigation focused on locating, identifying, and evaluating economically viable mineral deposits~\cite{dentith2024geophysics}. 
It is essential to discover and secure raw materials critical for global infrastructure, technological advancement, and sustainable development~\cite{gocht2012international}. 
Nowadays, remote-sensing satellite imaging studies are widely and effectively used in mineral exploration, offering an efficient, cost-effective alternative to traditional field surveys~\cite{Van2012Multi,Bedini2017use,Ousmanou2024Mapping}.

In remote-sensing mineral exploration, human experts typically follow: identifying \textit{geological} features from images like faults and fractures, integrating multiple \textit{hyperspectral} images to detect mineralization patterns, and synthesizing these into a mineral prospectivity map (MPM) to predict mineral deposit locations~\cite{Sabins1999Remote,mineral_review,mpm}. These processes are manual, time-intensive, and reliant on expert knowledge, necessitating automated machine learning and deep learning (DL) solutions for scalability.
% there are usually multiple remote-sensing images to depict the geological and hyperspectral features in a comprehensive way, and a widely-used workflow~\cite{Sabins1999Remote,mineral_review,mpm} on the images by a professional engineer involves three essential steps: 
% 1) identifying and examining geological features, such as faults and fractures, 
% 2) integrating hyperspectral data to detect various mineralization pattens, 
% and 3) synthesizing into a mineral prospectivity map (MPM), which reflects the likelihood and location of mineral deposits.
% Thus, this workflow is inherently complex, requiring advanced reasoning across multi-source RS data, combined with domain-specific knowledge.

To this end, DL models, e.g., CNNs~\cite{cnn1,cnn2} and Transformers~\cite{transformer1}, have been widely applied to process remote-sensing (RS) data. These methods excel at extracting features from RS images, mapping geological, geochemical, and geophysical attributes to predict mineral deposits~\cite{8}. However, their data-driven nature makes them reliant on labeled datasets, limiting their generalization to new geological scenarios. 
% To achieve this, various models, such as convolutional neural networks, and the Transformers, have been applied to process multi-source RS data~\cite{cnn1,cnn2,transformer1}.
% These approaches are particularly effective for analyzing RS data, as they can efficiently map the geological, geochemical, and geophysical data associated with the targeted mineral deposit type~\cite{8}.
% However, their effectiveness is constrained by the scarcity of labeled datasets and the inability to incorporate additional domain-specific textual information. 
% As a result, each region typically requires dozens of labeled data or additional expert analysis, which greatly limits the practical applicability.
In contrast, multimodal large language models (MLLMs) have recently emerged with remarkable zero-shot capabilities, integrating visual and textual contexts to tackle tasks~\cite{10,11,12} without requiring task-specific training data~\cite{9}. 

% MLLMs for rs
% Fortunately, Multi-modal Large Language Models (MLLMs) have recently emerged with remarkable zero-shot capabilities in integrating visual and textual information, effectively addressing the data-scarcity challenge~\cite{9}. MLLMs have demonstrated significant progress across remote-sensing evaluation tasks, including visual question answering, object detection, image segmentation, and location identification~\cite{10,11,12}. However, when provided with multi-image remote-sensing data and domain-specific instructions, MLLMs encounter significant interpretative challenges~~\cite{13,14}.

% experimental support
\begin{figure*}[tbp]
\centering
\includegraphics[width=0.87\linewidth, trim=0pt 30pt 0pt 0pt, clip]{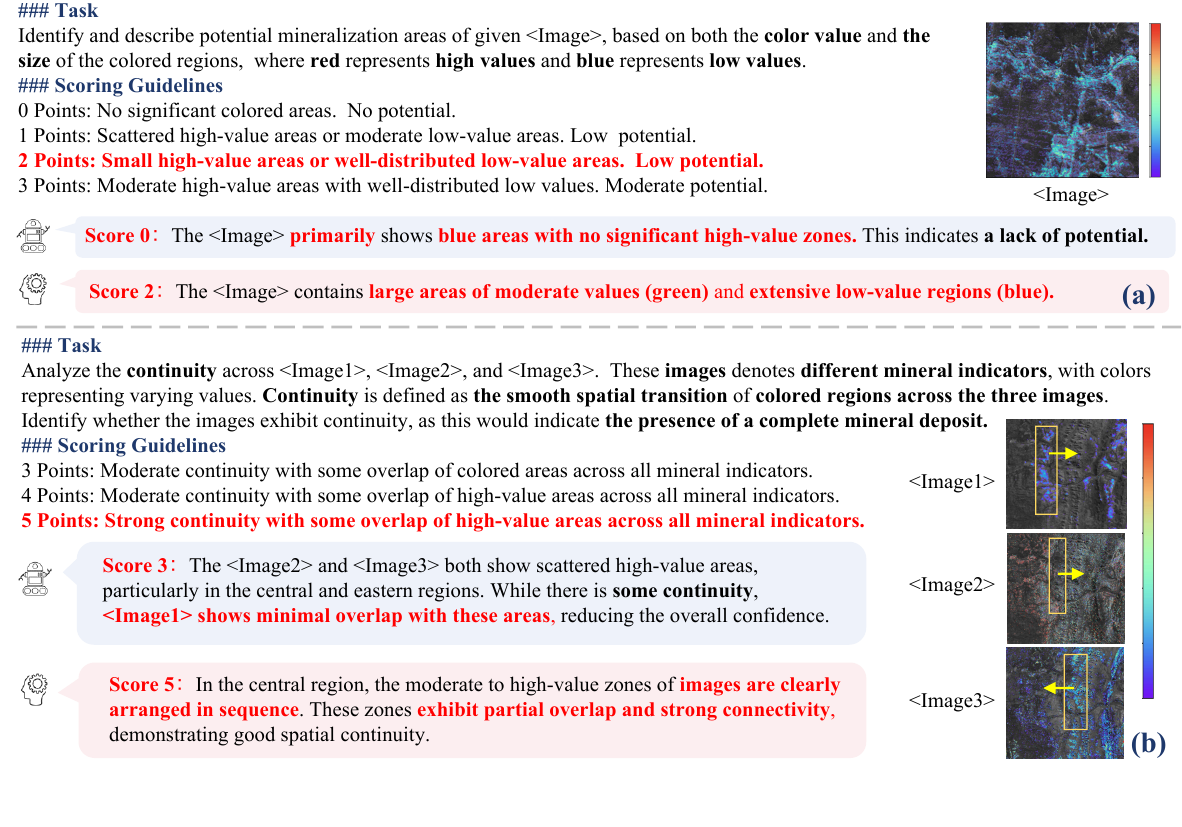}
\caption{\small
Judgment comparisons between GPT-4o~\cite{chatgpt4o} and human evaluator.
% GPT-4o~\cite{chatgpt4o} and a human evaluator were tasked with scoring based on a given prompt, scoring guidelines, and related images. 
\textcolor{darkblue}{GPT-4o} in blue box and \textcolor{darkred}{human-annotation} in red box.
% The blue box contains the description generated by \textcolor{darkblue}{GPT-4o}, while the \textcolor{darkred}{human-annotated} descriptions are shown in red. 
In (b), yellow boxes highlight regions and their spatial relations identified by the human but not GPT-4o.
% showing more advanced reasoning results not captured by MLLMs.
}
\label{fig:case}
\end{figure*}

Despite their promise, MLLMs face critical challenges when applied to mineral exploration with multiple RS images: they \textit{lack domain-specific geological knowledge} \citep{zhang2024mm} and \textit{struggle to reason effectively across multiple images} \citep{liu2024mibench,zhao2024benchmarking}. What's worse, domain-specific instructions (e.g., thousands of tokens) and multi-image inputs (e.g., 9 images) exacerbate the challenge by long-context issues that hinder reasoning accuracy \cite{liu2024lost}. 
For example, Figure \ref{fig:case}(a) highlights how MLLMs struggle with critical spatial elements like the strategic placement of low-value areas (blue). besides, in Figure \ref{fig:case}(b), MLLMs demonstrate significant difficulty in synthesizing spatial relationships across multiple geological contexts.
% These challenges are particularly pronounced in domain-specific image reasoning, as illustrated in Figure \ref{fig:case}(a).
% The disparity between the MLLMs' and human evaluators arises from the nuanced contextual understanding inherent in human perception, which current MLLMs fail to replicate.
% The MLLMs low score overlooks key factors, such as the strategic placement of low-value areas (blue), essential for interpreting scenes.
% The complexity intensifies when MLLMs are tasked with multi-image reasoning. 
% Moreover, Figure \ref{fig:case}(b) denotes the MLLM's difficulty in synthesizing spatial relationships and interconnections between disparate geological contexts.
% Consequently, MLLMs fail to assess mineral potential accurately due to these two primary limitations: \emph{domain-specific gaps in interpreting nuanced visual details and limitations in multi-image reasoning.}

\begin{figure*}[htb]
    \centering\includegraphics[width=0.85\linewidth, trim=0pt 315pt 220pt 0pt, clip]{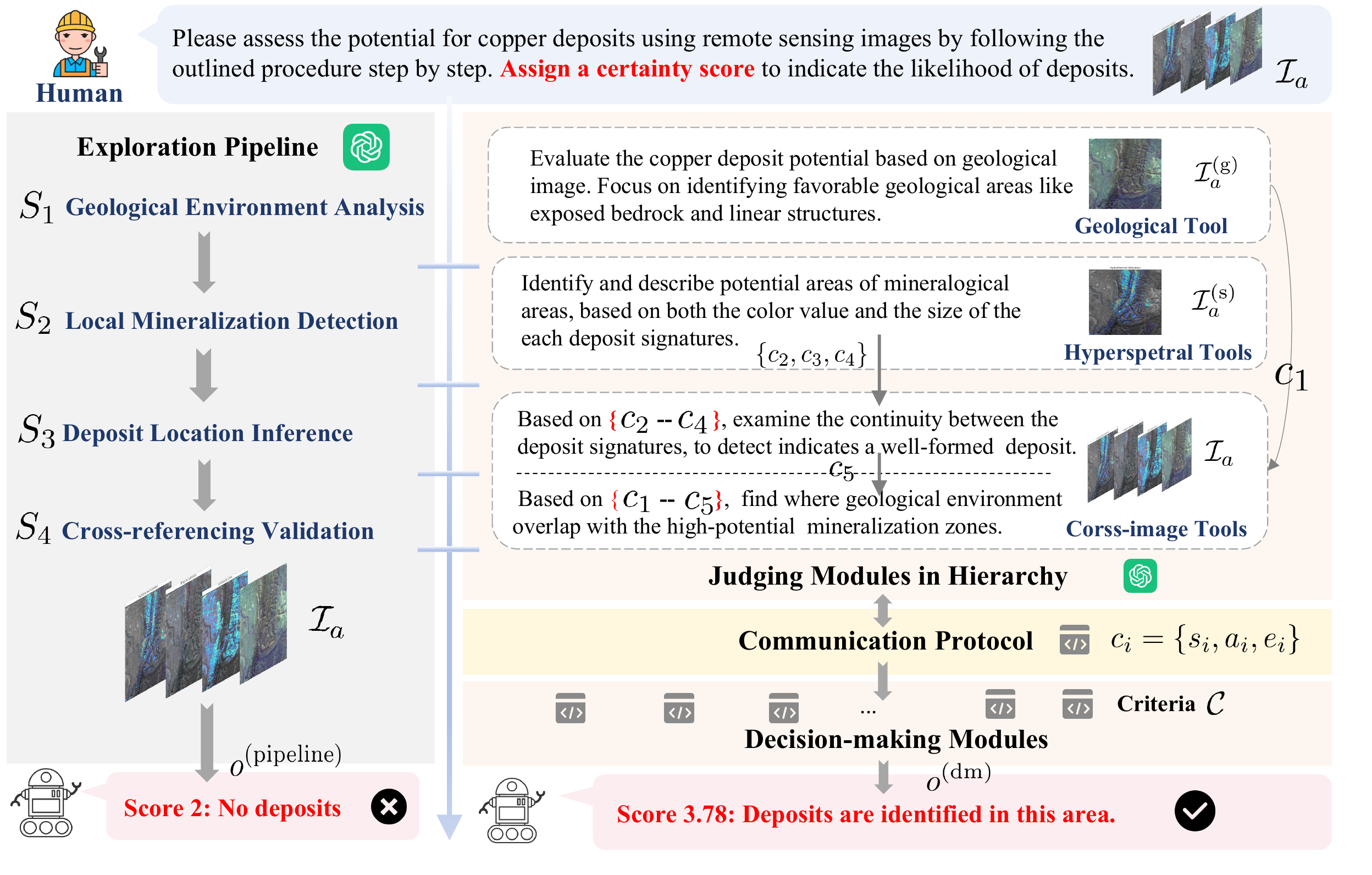}
    \caption{The tailored MineAgent for mineral exploration. (Left) Base pipeline using step-by-step reasoning; (Right) MineAgent decomposing pipeline into specialized modules, improving assessment accuracy.}
    \label{fig:frame}
\end{figure*}

To circumvent the challenges of domain-specific knowledge integration and multi-image reasoning in mineral exploration, we propose MineAgent, a modular MLLM framework specifically designed to address these complexities (Figure~\ref{fig:frame}). MineAgent employs hierarchical judging and decision-making modules to extract, integrate, and analyze spatial and spectral features from remote-sensing data. By considering the common-practice pipeline and orchestrating these modules, MineAgent enhances reasoning capabilities across geological and hyperspectral contexts, enabling accurate deposit predictions.
% To address existing limitations, we present MineAgent, a specialized MLLM agent designed to replicate expert mineral exploration techniques, as shown in Figure \ref{fig:frame}. 
% Built upon MLLMs, MineAgent demonstrates unprecedented capabilities in single- and multi-image interpretation without requiring additional training data.
% Further, MineAgent integrates a local-global visual reasoning module, enabling sophisticated multi-image reasoning.
% Additionally, we developed a collaborative evaluation with quantitative assessment capabilities to enhance domain-specific knowledge comprehension and enable error-tailored reasoning.
% This modular design significantly enhances the accuracy and reliability of mineral potential assessment. 

Furthermore, we present MineBench, a benchmark tailored for multimodal mineral exploration tasks. MineBench standardizes task formulations and datasets, enabling rigorous evaluation of MLLMs in reasoning over domain-specific remote-sensing data. 
MineBench provides a unique challenge to MLLMs where both multi-image reasoning and long-tail domain understanding are required to accomplish one task. 
% Another critical gap in mineral exploration has been the lack of standardized benchmarks for performance evaluation. Thus, we developed MineBench, a mineral exploration benchmark spanning 612 distinct geological regions. 
% Using MineBench, we systematically evaluate proprietary and open-source MLLMs, providing unprecedented insights into their capabilities and limitations. 
% Moreover, the dataset is strategically stratified into three difficulty levels, enabling nuanced and rigorous assessment of MLLMs. 

The main contributions of this work are:
\vspace{-5pt}
\begin{itemize}
\setlength{\itemsep}{0pt}
\setlength{\parskip}{3pt}
\item We propose MineAgent, a modular framework addressing domain challenges in multi-image reasoning for mineral exploration.
\item We develop MineBench, a standardized benchmark for evaluating MLLMs in mineral exploration with geological and hyperspectral data.
\item We conduct extensive experiments across various models and setups, demonstrating MineAgent's effectiveness and providing insights into MLLMs in this domain.
\end{itemize}

\section{Related Work}

\paragraph{Multi-image Reasoning of MLLMs.}
Recent studies have revealed a significant performance gap between single- and multi-image reasoning tasks~\cite{13,wang2024muirbench,jiang2024mantis, zhang2024cocot, liu2024mibench}. 
For instance, \citet{14} highlighted deficiencies in temporal and contextual reasoning across image sequences while \citet{zhao2024benchmarking} explored multi-dimensional aspects of multi-image reasoning, such as perception, knowledge integration, reasoning, and multi-hop inference.
However, domain-specific tasks, such as those in mineral exploration, pose unique challenges requiring not only multi-image reasoning but also domain-specific knowledge. This motivates us to present MineBench to evaluate MLLMs' reasoning capabilities within a long-tail domain rigorously.

% Despite these advancements, a critical gap persists in enabling nuanced multi-image reasoning within domain-specific contexts. To address these challenges, MineBench includes targeted tasks to test MLLMs' capabilities in domain-specific reasoning, offering a pathway for advancing their applicability in specialized fields.

\paragraph{Remote Sensing MLLMs.}
The application of MLLMs in remote sensing has gained traction for tasks like image captioning and visual question answering~\cite{zhan2023rsvg,cheng2022nwpu,wang2021loveda}. These models, fine-tuned with extensive visual-text instructions, demonstrate strong performance on single-image tasks~\cite{kuckreja2024geochat,zhan2024skyeyegpt,luo2024skysensegpt}. However, their capabilities remain limited when extended to multi-image reasoning, a critical requirement for mineral exploration tasks that demand integration of spatial and spectral information across multiple images.
Furthermore, the lack of standardized datasets tailored to multi-image remote-sensing tasks hinders progress in applying MLLMs to domains like mineral exploration. Addressing this gap, we propose a modular MLLM framework, MineAgent, coupled with MineBench.
% , a benchmark designed to evaluate and enhance multi-image reasoning for domain-specific challenges.
% The application of MLLMs in remote sensing has grown rapidly, encompassing tasks such as image captioning and visual question answering~\cite{zhan2023rsvg,cheng2022nwpu,wang2021loveda}. 
% Fine-tuned on extensive visual-text instructions, these models exhibit strong performance~\cite{kuckreja2024geochat,zhan2024skyeyegpt,luo2024skysensegpt}.
% However, most existing approaches are limited to single-image inputs, significantly restricting their effectiveness in multi-image reasoning applications.
% This limitation is particularly pronounced in mineral exploration, where the task inherently requires multiple images to analyze mineralization patterns~\cite{transformer1}. Compounding the challenge is the lack of standardized datasets, which limits the scalability of remote-sensing MLLMs. 
% Recognizing this gap, we introduce an innovative MLLM agent framework tailored to address these multi-image reasoning challenges. By extending the capabilities of MLLMs to integrate and interpret multi-image data, this framework paves the way for their application in complex remote-sensing tasks.

\section{MineBench: on Remote-sensing Images}

The field of mineral exploration currently lacks a well-organized benchmark to evaluate the performance of MLLMs. Existing ones do not capture the unique challenges of integrating geological knowledge with multimodal reasoning. These motivate us to present a new mineral exploration benchmark.

\paragraph{Task Formulation.} 
Mineral exploration enables quantifiable assessments of deposit likelihood, facilitating prioritization of exploration areas, so it is usually formulated as a binary classification problem~\cite{cnn1,cnn2}, i.e.,
\begin{align}
% \vspace{-5pt}
    y \sim P_{\theta}(\ry|a) \coloneqq P_{\theta}(\cdot|\gI_a) = P_{\theta}(\cdot|\gI_a^{\text{(g)}}, \gI_a^{\text{(h)}})
\end{align}
where $y\in\gY$ represents the presence of mineral deposits with $\gY=\{0, 1\}$ in a targeted area $a$ and $a$ can be represented as a set of remote-sensing images, $\gI^a$. 
In mineral exploration, according to distinct roles and nature of the data, $\gI^a$ can be coarsely grouped into two sub-sets, i.e., geological images ($\gI_a^{\text{(g)}}$) and hyperspectral images ($\gI_a^{\text{(h)}}$).

\begin{figure}[tbp]
    \includegraphics[width=1\linewidth, trim=0pt 200pt 20pt 0pt, clip]{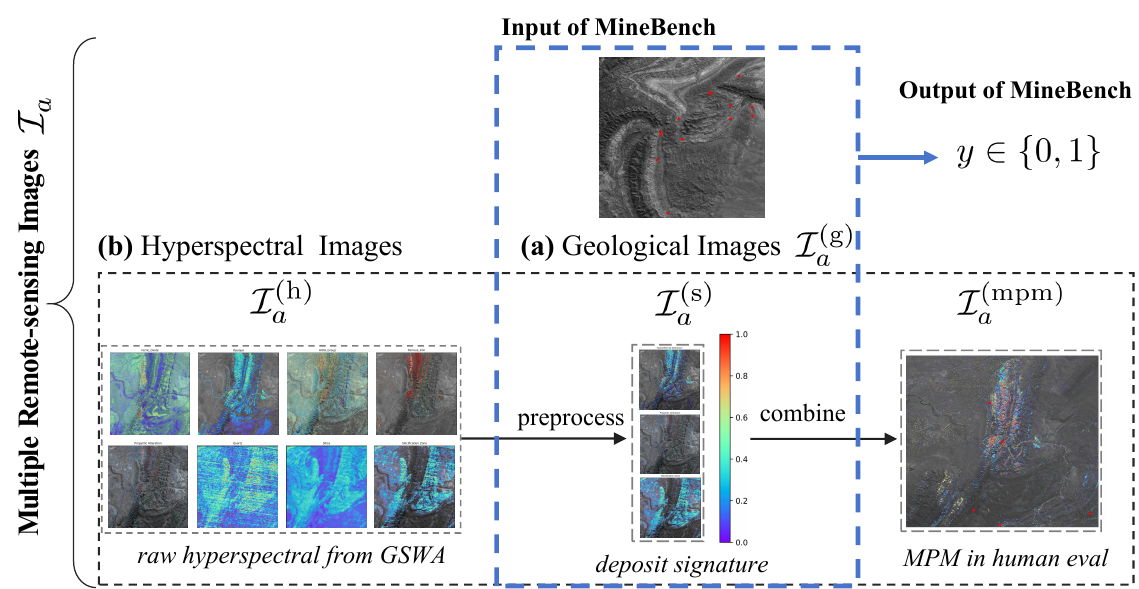}
    \caption{
    \small
    Task definition in MineBench. 
    Particularly, a targeted area $a$ is represented by two image types, i.e., $\gI_a=\{\gI_a^{\text{(g)}}, \gI_a^{\text{(h)}}\}$.  $\gI_a^{\text{(h)}}$ are color-coded images where uncolored regions represent sub-threshold potential. 
    }
    \label{fig:data}
\end{figure}

\paragraph{Remote-sensing Images in Mineral Exploration.} 
\textit{Geological images} ($\gI_a^{\text{(g)}}$, see Figure~\ref{fig:data}(a)) focus on capturing macroscopic spatial and morphological features, such as landforms, tectonic structures, and geological units, which are critical for identifying large-scale mineralization patterns and structural controls, whereas \textit{hyperspectral images} ($\gI_a^{\text{(h)}}$, see Figure~\ref{fig:data}(b)) are designed to capture high-resolution spectral reflectance data across a wide range of wavelengths, enabling the detailed characterization and differentiation of mineral types and their spatial distributions at a pixel-by-pixel level. 

\paragraph{Data Sourcing and Preprocessing.} 
We utilized raw remote-sensing data from the Geoscience Western Australia (GSWA) repository\footnote{\url{https://data.dea.ga.gov.au/?prefix=ASTER_Geoscience_Map_of_Australia/}}, an open-source geoscience data source~\cite{dataset}, to compose MineBench. 
Although the raw remote-sensing images encompass all necessary information to infer mineral exploration tasks, the raw hyperspectral images, $\gI_a^{\text{(h)}}$, are not intuitive for visualizing mineralization patterns~\cite{Sabins1999Remote}. This limitation makes it considerably challenging for MLLMs or even humans to identify meaningful deposit signatures directly because such tasks require extensive domain knowledge in mineral exploration to interpret and process the raw data effectively. Therefore, following common practices in mineral exploration~\cite{alteration,wambo2020identifying, ghamisi2017advances}, we preprocess the raw hyperspectral images using domain-specific linear combinations to generate three distinct deposit signature images, $\gI_a^{\text{(s)}}$ (see Figure~\ref{fig:data}(bottom)). These signature images provide a visually interpretable representation of key features in deposit formation, significantly reducing task complexity for both human experts and automated models. Consequently, MineBench is formulated as
% \vspace{-5pt}
\begin{align}
    y \sim P_{\theta}(\ry|\gI_a^{\text{(g)}}, \gI_a^{\text{(s)}}).
\end{align}
More details on data sourcing and preprocessing are provided in Appendix~\ref{sec:construction} and \ref{sec:preprocess}, respectively.

\paragraph{Data Labeling and Sampling.} 
We access the deposit locations directly sourced from authoritative records as class labels\footnote{\url{https://map.sarig.sa.gov.au/}} -- `positive' as $y = 1$ and `negative' as $y = 0$.
Considering the inherent class imbalance in real-world mineral exploration, a strategic random sampling approach was employed to ensure robust evaluation close to real-world scenarios. 
The resulting dataset consists of 73 positive areas and 539 negative areas, yielding an $\sim$1:9 positive-to-negative ratio, leading to reliable and statistically meaningful evaluations~\cite{15,16}.

\paragraph{Validation of Preprocssing.}
While preprocessing provides visually interpretable features for mineral exploration, it inherently involves a loss of information due to the linear combinations. To ensure that the processed data retains sufficient detail for deterministic judgments, we also conducted a human evaluation on a subset (20\%) of the dataset. 
To facilitate this evaluation, the deposit signature images, $\gI_a^{\text{(s)}}$, were further combined into a mineral prospectivity map (MPM), $\gI_a^{\text{(mpm)}}$, offering a clear and intuitive visualization of potential mineral deposits ~\cite{zuo2020geodata, xu2021mineral}. 
Using the MPM, human experts make judgments upon
% \vspace{-5pt}
\begin{align}
    \text{HumanEval}(\cdot|\gI_a^{\text{(g)}}, \gI_a^{\text{(mpm)}})
\end{align}
and validated the data by comparing it against official deposit locations. The results (97.4\% accuracy) demonstrate that even with MPM, human judgments align well with the provided class labels, confirming the reliability of the preprocessing. Further details on the validation process are provided in Appendix~\ref{sec:validation}. 
In the remaining, we omit the subscript $a$ for clarity if no confusion is caused.

\section{Methodology}

To leverage the capabilities of multimodal large language models (MLLMs) in mineral exploration, we naturally formulate the task as a visual question-answering (VQA) problem. 
Specifically, given remote-sensing images (e.g., $\gI^{\text{(g)}}$ and $\gI^{\text{(s)}}$) and a domain-specific query about the presence of a particular mineral deposit, the model generates a response indicating the likelihood of the deposit. This response can then be mapped to a classification label using a predefined verbalizer, i.e.,
\begin{align}
    o \sim \gM(\gI^{\text{(g)}}, \gI^{\text{(s)}};\theta),\label{equ:mllm-task-form}
\end{align}
where $\theta$ parameterizes the MLLM, $\gM$, and $o$ denotes a natural language response with verbalizer.

Despite the simplicity in Eq.(\ref{equ:mllm-task-form}), the inherent limitations of MLLMs in lacking domain-specific geological knowledge pose significant challenges -- they struggle with interpreting complex hyperspectral imaging data and understanding spatial patterns critical for mineral exploration.

\subsection{Baseline: Mineral Exploration Pipeline}
\label{4.1}
To alleviate the lack of domain knowledge, we first propose a baseline framework that integrates the conventional mineral exploration pipeline with domain-specific instructions to enhance the understanding and reasoning capabilities of MLLMs. 
This pipeline emulates the workflow of human experts in mineral exploration by leveraging step-by-step structured prompts and reasoning mechanisms.

Formally, let $P$ denote a curated set of domain-specific prompts tailored for the task. These prompts are carefully designed to encode key domain knowledge and guide the MLLM through sequential reasoning steps. 
The method can be represented as
\begin{align}
    o^{\text{(pipeline)}} \sim \gM(\gI^{\text{(g)}}, \gI^{\text{(s)}};\theta, P),
\end{align}
where $\gI^{\text{(g)}}$ and $\gI^{\text{(s)}}$ represent geological and processed hyperspectral images, and $o^{\text{(pipeline)}}$ is the model-generated step-by-step reasoning by following the pipeline instruction $P$. 

Specifically, pipeline instruction $P$ 
consists of sequential stages that transform raw geological data into actionable insights:
\emph{Geological Environment Analysis ($S_1$)}
analyzes geological images to identify key structural elements, such as faults, crucial for understanding the mineralization environment.
\emph{Local Mineralization Detection ($S_2$)} 
uses hyperspectral images to detect deposit signatures by examining color variations, providing granular insights into potential mineralization zones.
\emph{Global Deposit Location Inference ($S_3$)}
evaluates spatial correlations among deposit signatures to infer deposit locations based on mineralization patterns. Sequential arrangements with strong spatial continuity suggest the presence of complete deposits.
\emph{Cross-referencing Validation ($S_4$)} 
synthesizes findings from $S_1$ to $S_3$ to estimate deposit probabilities and accurately identify target exploration areas.
Therefore, $o^{\text{(pipeline)}}$ can be decomposed as
\begin{align}
    o^{\text{(pipeline)}} \coloneqq o^{\text{(s1)}}\oplus o^{\text{(s2)}}\oplus o^{\text{(s3)}}\oplus o^{\text{(s4)}}\oplus y,
\end{align}
where $o^{\text{(s*)}}$ denotes the rationale and staged judgment generated for the corresponding stage, and $y$ is the final judgment. 
Note we generate all outputs in $o^{\text{(pipeline)}}$ together in one MLLM inference (see Appendix~\ref{sec:case-study} for details). 
As such, this structured pipeline enhances the reliability of mineral exploration by ensuring transparency at every stage.

% \subsection{MineAgent: }

% modules
% - categorical image understanding
%   - geographical understanding
%   - hyperspectral understanding
% - Task-oriented reasoning
% - judging and xxx

\begin{table*}[t]
    \centering
    \scriptsize
    \renewcommand{\arraystretch}{1.5}
    \begin{tabular}{l c l c c c c}
    \hline
        \textbf{Tool Type} & \textbf{Module Type} & \textbf{Captured Feature}  & \textbf{Inp. Imgs ($\gI$)}    & \textbf{Inp. Ref ($\gC$)}  & \textbf{Output}     & \textbf{Stage}  \\ \hline
        Geological Tool & Judging  & Geological context  & $\gI_a^{\text{(g)}}$    & N/A & $c_1$   & $S_1$   \\  \hline
        \multirow{3}{*}{Hyperspetral Tools} & Judging  & Signature 1 & $\gI^{\text{(s)}}$    & N/A & $c_2$          & $S_2$     \\ 
        ~     & Judging & Signature 2 & $\gI_a^{\text{(g)}}$   & N/A & $c_3$      & $S_2$     \\ 
        ~      & Judging & Signature 3 & $\gI_a^{\text{(s)}}$   & N/A & $c_4$    & $S_2$    \\ \hline
        \multirow{2}{*}{Corss-image Tools}   & Judging  & Relation between signatures & $\gI^{\text{(s)}}$       &  $c_2$--$c_4$     & $c_5$       & $S_3$  \\  
    ~       & Judging     & Validation between $\gI_a^{\text{(g)}}$ and $\gI_a^{\text{(s)}}$    & $\gI_a^{\text{(g)}}$, $\gI_a^{\text{(s)}}$ &  
    $c_1$--$c_5$ & $c_6$ & $S_4$ \\ \hline
    Deposit Presence Tool    & Decision-making    & Response of deposit presence       & N/A       & N/A       & $o^{\text{(dm)}}$  & N/A \\ \hline
    \end{tabular}
    \caption{\small Tools in MineAgent tailored for mineral exploration. `Inp.' denotes the model inputs in Eq.(\ref{equ:judging_modules}).}
    \label{tab:tools_summary}
\end{table*}

\subsection{MineAgent: Orchestrating over Remote-sensing Images}
\label{4.2}
However, the above is still vulnerable to multi-image reasoning as MLLMs struggle to synthesize spatial and contextual relationships across multiple images and long contexts, leading to incomplete understandings of multiple remote-sensing images for mineral exploration. Therefore, we propose an agent framework, MineAgent, that decomposes the exploration process into modular components.
% , allowing for parallel reasoning and robust integration of intermediate results. 

\subsubsection{General MineAgent Framework}
\begin{figure}[t]
    \includegraphics[width=0.88\linewidth, trim=0pt 310pt 230pt 0pt, clip]{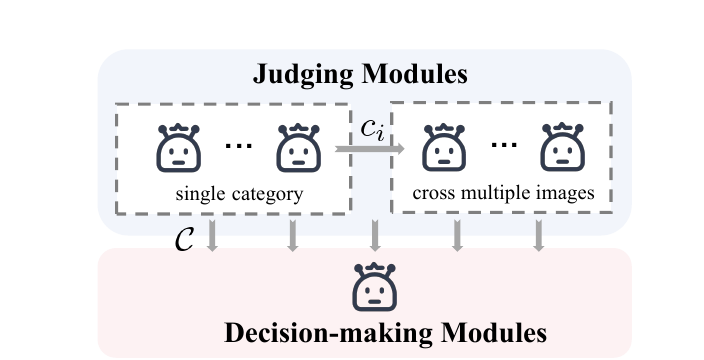}
    \caption{
    \small
    A general framework of MineAgent. }
    \label{fig:agent}
\end{figure}

MineAgent focuses on two core kinds of modules to enhance its reasoning capabilities: \textit{judging} and \textit{decision-making}, as shown in Figure~\ref{fig:agent}. 
While a judging module specializes in extracting specific features from remote-sensing images (e.g., geological structures or spectral mineralization signatures), a decision-making module is designed for a specific task to integrate these features to perform high-level reasoning tasks, such as inferring mineral deposit likelihood or validating predictions across diverse datasets. 

% While scoring modules evaluate agent actions in multi-image reasoning scenarios, the judgment module orchestrates workflows and decision-making by determining the sequence and conditions under which these evaluations occur.

\paragraph{Judging Modules in Hierarchy.}
A judging module operates with two types of inputs to maintain focus and robustness in feature extraction, i.e., 
\begin{align}
    c \sim \gJ(\gI, \gC;\theta),~\text{where}~\gC=\{c_i\}_{i=0}^N.\label{equ:judging_modules}
\end{align}
The first type of input includes one or a few single-category remote-sensing images, $\gI$, such as geological or hyperspectral images, ensuring that the module specializes in analyzing a specific feature (e.g., structural patterns or deposit signatures). 
The second type of input comes from the outputs, $\gC$, of other judging modules, enabling a hierarchical structure. And either of them is optional. 
This setup allows for cross-image reasoning and intermediate result validation, effectively integrating insights from multiple sources. By focusing on specific features while facilitating inter-module communication, this approach circumvents the challenges of multi-image reasoning in complex mineral exploration tasks. 

\paragraph{Communication Protocol.} 
A well-defined communication protocol is critical for ensuring efficient information exchange between modules. Each module in MineAgent communicates using a semi-structured output format,
\begin{align}
\label{equ:output_format}
  c_i = \left\{ s_i, a_i, e_i \right\},
\end{align}
where $s_i$  is a numerical score reflecting the module's confidence or evaluation, $a_i$  represents the identified favorable areas or features, and $e_i$ provides an analytical explanation or rationale behind the module's output. $s_i$ is determined based on a detailed scoring guide, ensuring consistency.

% The scoring module employs a hierarchical flow paradigm to address the complex challenges of multi-image reasoning. 
% Initially, domain-specific modules $m^{\text{(domain)}}$ are used to analyze single-image features, and these outputs $o^{\text{(domain)}} = \{o_1^{\text{(single)}},\cdots, o_i^{\text{(single)}}\} $ are synthesized by cross-referencing modules $m^{\text{(cross)}}$ to infer relationships across multiple images, represented as:
% \begin{align}
%     o^{\text{(cross)}}  \sim \gM(\gI^{a};\theta, P; o^{\text{(domain)}})
% \end{align}
% where the outputs $o^{\text{(cross)}}$ may includes various results as $o^{\text{(cross)}} = \{o_1^{\text{(multi)}},\cdots, o_k^{\text{(multi)}}\}$
% This hierarchical flow paradigm facilitates a comprehensive understanding of both detailed features in single images and their broader interconnections.
% To support inter-agent communication, the scoring module includes an output protocol tailored to application-specific requirements, thereby promoting seamless message-passing during reasoning tasks.
% Extending its analytical capabilities, the scoring module integrates visual-textual scoring criteria, enabling precise evaluations by assigning distinct scores.

\paragraph{Decision-making Modules Specific to Tasks. }
A decision-making module is invoked to integrate multiple structured outputs from the judging modules to derive high-level insights and outputs for a specific task. Formally, 
this task-specific module is written as
\begin{align}
    o^{\text{(dm)}} \coloneqq \gD(\gC; \theta), \label{equ:decisionmaking_modules}
    % o_1^{\text{(single)}}\oplus \cdots \oplus o_{i+k}^{\text{(multi)}}\oplus y,
\end{align}
The module takes a set of assessment tuples $C = \{c_1, c_2, \ldots, c_M\}$ as input and outputs $o^{\text{(dm)}}$, the final decision, along with a confidence score or feedback to guide downstream processes.

Thus, MineAgent orchestrates the exploration process by integrating judgments from hierarchical judging modules and decisions from task-specific decision-making modules, ensuring robust multi-image reasoning and high-confidence answers.

% The judgment module is integral to orchestrating the overall workflow, and effectively managing pipelines by processing textual and visual information.
% The module breaks the pipeline into atomic tasks within the scoring module, which can be executed sequentially or in parallel depending on the query's complexity. Crucially, the judgment module receives outputs 
% from the scoring module and makes decisions by iterating over actions in the scoring agent. 
% If intermediate outputs fail to meet the defined protocol, this occurs within a pre-specified reasoning limit.
% This iterative and rigorous process ensures thorough and reliable reasoning.
% Upon comprehensive validation, the judgment module aggregates single- and multi-image outputs $\{ o_1^{\text{(single)}},\cdots,o_i^{\text{(single)}}, o_{i+k}^{\text{(multi)}}\}$ across the decomposed pipeline to generate the final decision, represented as
% \begin{align}
%     o^{\text{(judgment)}} \coloneqq o_1^{\text{(single)}}\oplus \cdots \oplus o_{i+k}^{\text{(multi)}}\oplus y,
% \end{align}
% By reusing or extending built-in capabilities, developers can configure agents for a wide range of tasks: writing code, executing computations, incorporating human feedback, validating outputs, or addressing specific multi-image reasoning challenges.

\subsubsection{Grounding for Mineral Exploration}
As we have a mature pipeline for mineral exploration with remote-sensing images according to human experts, we apply a workflow-based paradigm~\citep{li2024autoflow} to our agent framework for mineral exploration.

To ground the MineAgent, we propose multiple tools as in Table~\ref{tab:tools_summary} for judging modules in Eq.(\ref{equ:judging_modules}) and decision-making modules in Eq.(\ref{equ:decisionmaking_modules}): 
\emph{remote-sensing image judging tool} suite is a collection of MLLM-based modules designed to extract critical features from remote-sensing data, including geological and hyperspectral images (see Appendix~\ref{sec:case-study} for details). 
And \emph{deposit presence decision-making tool} insights from the judging modules to determine the likelihood of mineral deposit presence in a targeted area. To optimize computational efficiency, we directly employ Bayesian optimization~\cite{bayesian} to calculate the overall evaluation score $o^{\text{(dm)}} = \sum \text{w}_i s_i$, where $w_i$ represents the weight assigned to each criterion.

\begin{table*}[t]
\centering
\scriptsize
\renewcommand{\arraystretch}{1.3} % Adjust row height for better readability
\setlength{\tabcolsep}{5pt} % Adjust column spacing
\begin{tabular}{l l llll llll} 
\specialrule{1pt}{0pt}{2pt} % Thick top line
\multirow{3}{*}{\textbf{Source}} & \multirow{3}{*}{\textbf{Model}} & \multicolumn{4}{c}{\textbf{Baseline}} & \multicolumn{4}{c}{\textbf{With MineAgent}} \\ 
\cmidrule(r){3-6} \cmidrule(l){7-10} 
 & & \textbf{Pos.F1} & \textbf{Avg.F1} & \textbf{Roc-AUC} & \textbf{MCC} 
 & \textbf{Pos.F1} & \textbf{Avg.F1} & \textbf{Roc-AUC} & \textbf{MCC} \\ 
\specialrule{0.4pt}{2pt}{2pt} % Medium-thick line separating headers
\textbf{N/A} & Random Choice & 11.86 & 49.96 & 51.01 & 2.01 & 11.86 & 49.96 & 51.01 & 2.01 \\ 
\specialrule{0.4pt}{1.2pt}{1.2pt} % Medium-thick line separating sections
\multirow{4}{*}{\textbf{Closed-Source}} 
& Gemini-Pro-1.5 & 20.95 & 18.56 & 49.66 & -0.77 & 21.72 (\textbf{\textcolor{darkred}{+0.77}}) & 30.57 (\textbf{\textcolor{darkred}{+12.01}}) & 52.34 (\textbf{\textcolor{darkred}{+2.68}}) & 3.52 (\textbf{\textcolor{darkred}{+4.29}}) \\ 
& Gemini-Flash-2.0 & 20.30 & 41.24 & 51.33 & 1.73 & 22.54 (\textbf{\textcolor{darkred}{+2.24}}) & 56.18 (\textbf{\textcolor{darkred}{+14.49}}) & 56.03 (\textbf{\textcolor{darkred}{+4.70}}) & 12.37 (\textbf{\textcolor{darkred}{+10.6}}) \\ 
& Qwen2-VL-72B & 22.97 & 20.76 & 54.44 & 9.97 & 58.71 (\textbf{\textcolor{darkred}{+34.74}}) & 75.30 (\textbf{\textcolor{darkred}{+54.54}}) & 84.01 (\textbf{\textcolor{darkred}{+29.6}}) & 54.21 (\textbf{\textcolor{darkred}{+41.84}}) \\ 
\rowcolor[HTML]{FDEFF1} 
&  GPT-4o & 34.93 & 57.27 & 69.35 & 26.16 
& 61.20 (\textbf{\textcolor{darkred}{+26.27}}) & 77.19 (\textbf{\textcolor{darkred}{+19.92}}) & 83.35 (\textbf{\textcolor{darkred}{+13.82}}) &  56.30 (\textbf{\textcolor{darkred}{+30.14}}) \\ 
\specialrule{0.4pt}{1.2pt}{1.2pt} % Medium-thick line separating sections
\multirow{5}{*}{\textbf{Open-Source}} 
% & LLaVA-7B & 21.39 & 27.44 & 51.39 & 2.25 
% & 21.26 (\textbf{\textcolor{darkblue}{-0.13}}) & 22.41 (\textbf{\textcolor{darkblue}{-5.03}}) & 50.62 (\textbf{\textcolor{darkblue}{-0.77}}) & 1.18 (\textbf{\textcolor{darkblue}{-1.07}}) \\ 
& Yi-6B & 15.74 & 43.89 & 47.14 & -3.82 
& 21.82 (\textbf{\textcolor{darkred}{+6.08}}) & 15.16 (\textbf{\textcolor{darkblue}{-28.74}}) & 51.54 (\textbf{\textcolor{darkred}{+7.65}}) & 5.05 (\textbf{\textcolor{darkred}{+8.87}}) \\ 
& LLaVA-13B & 19.46 & 26.32 & 47.28 & -4.31 
& 20.58 (\textbf{\textcolor{darkred}{+1.9}})  & 21.36 (\textbf{\textcolor{darkblue}{-4.96}}) & 50.77 (\textbf{\textcolor{darkred}{+3.49}})& 1.60 (\textbf{\textcolor{darkred}{+5.91}}) \\ 
& InterVL-26B & 19.26 &  31.87 & 56.91 & 8.95 
& 24.23 (\textbf{\textcolor{darkred}{+5.06}}) & 44.32 (\textbf{\textcolor{darkred}{+12.45}}) & 57.10 (\textbf{\textcolor{darkred}{+0.19}})& 9.24 (\textbf{\textcolor{darkred}{+0.29}}) \\ 
& Yi-34B & 22.16 & 43.63 & 54.28 & 5.55 & 29.71 (\textbf{\textcolor{darkred}{+7.10}}) & 50.85 (\textbf{\textcolor{darkred}{+7.22}}) & 64.82 (\textbf{\textcolor{darkred}{+10.34}}) & 19.28 (\textbf{\textcolor{darkred}{+13.73}}) \\ 
\rowcolor[HTML]{FDEFF1}  
&Qwen-7B & 22.92 & 24.16 & 54.59 & 8.69 
& 30.99 (\textbf{\textcolor{darkred}{+8.07}}) & 47.93 (\textbf{\textcolor{darkred}{+23.77}}) & 68.23 (\textbf{\textcolor{darkred}{+13.64}}) & 23.79 (\textbf{\textcolor{darkred}{+15.10}}) \\ 
\specialrule{1pt}{3pt}{0pt} % Thick bottom line
\end{tabular}
\caption{\small Performance comparison of models with and without MineAgent. Highlighted rows indicate the highest scores among closed-source and open-source MLLMs. \textcolor{darkred}{Red} represents improvements, while \textcolor{darkblue}{blue} denotes reductions.}
\label{table:performance}
\end{table*}

\section{Experiment}
\paragraph{Metric.}
Due to the inherent class imbalance in MineBench, multiple complementary metrics are employed. The F1 score for positive classes (Pos.F1) evaluates the MLLMs' ability to identify deposits. 
The macro-averaged F1 score (Avg.F1) provides balanced assessment across classes, while the ROC-AUC evaluates discriminative ability. 
Additionally, the Matthews Correlation Coefficient (MCC) provides a comprehensive evaluation by synthesizing the confusion matrix ranging from -1 to 1, where -1 indicates complete misclassification, 1 represents perfect prediction~\cite{mcc}.
The details of the experimental setup are provided in Appendix~\ref{sec:expriment_setup}.

\subsection{Main Results}
\label{main_results}
Table \ref{table:performance} presents the comparative performance of various MLLMs: Qwen2-VL-7B/72B~\cite{Qwen-VL,Qwen2VL}, Gemini-Flash-2.0~\cite{team2024gemini}, Gemini-Pro-1.5, GPT- 4o~\cite{chatgpt4o}, LLaVA-13B~\cite{liu2023llava,liu2024llavanext}, Yi-6B/34B~\cite{yi} and InterVL-26B~\cite{chen2024internvl} on MineBench. 
This evaluation highlights several key findings regarding the strengths and limitations of the MLLMs:

\paragraph{Effectiveness of MineAgent.}  
Experimental results demonstrate significant performance improvements achieved by MineAgent, with the highest improvement reaching 30.14\% when paired with GPT-4o and 23.77\% when paired with Qwen-7B.
This result highlights the potential of MineAgent in enhancing multi-image reasoning and domain-specific gaps. 
Few open-source MLLMs show performance degradation when integrated with MineAgent, primarily due to their unstable reasoning capabilities in interpreting score criteria. 
During base pipeline evaluations, inconsistent outputs also emerged, such as assigning a score of 5 as ``positive'' and a score of 9 as ``negative.''
Notably, Yi-6B exhibited a 23.53\% label-score mismatch,  while LLaVA-13B showed an even higher rate of 35.15\%, compared to just 1.96\% for GPT-4o. 
% The complexity intensifies with MineAgent's detailed visual-text criteria, which pose additional challenges for these MLLMs to comprehend accurately. 
These findings underscore the critical importance of stable foundational models in achieving consistent performance improvements.

\paragraph{Performance Ceiling.} 
MLLMs encounter significant limitations when addressing mineral exploration tasks, even with the enhanced reasoning capabilities provided by MineAgent. 
For instance, GPT-4o achieves a Pos.F1 score of only 61.21\% and an Avg.F1 of 77.19\%. 
Notably, several open-source MLLMs perform below the random choice, underscoring fundamental architectural constraints.
Moreover, a substantial performance disparity exists between closed-source and open-source MLLMs. 
% Closed-source MLLMs achieve peak Roc-AUC scores of 83.35\%, whereas open-source MLLMs exhibit more pronounced limitations, with peak MCC scores reaching only 68.23\%.
This performance gap stems from two factors: a lack of high-quality, domain-specific training data to capture the nuances and insufficient exposure to multi-image reasoning scenarios needed for handling task complexity.

% \textbf{(3) Performance Disparity Analysis:} 

% Open-source MLLMs demonstrate even more pronounced limitations, with peak Avg.F1 scores of 62.53\%, 47.93\%, and 15.78\% on ``Easy,'' ``Medium,'' and ``Hard'' settings, respectively.

\subsection{MLLM Capabilities Evaluation}
\begin{table}[b]
\centering
\small
\renewcommand{\arraystretch}{1.5} % Adjust row height for better readability
\setlength{\tabcolsep}{6pt} % Adjust column spacing
\begin{tabular}{c  cc  c}
\specialrule{0.8pt}{0pt}{2pt} % Thick top line
\multirow{2}{*}{\textbf{Setting}} & \multicolumn{2}{c}{\textbf{Input}} & \multirow{2}{*}{\textbf{Output}} 
\\ \cline{2-3}
~ & Inp.Imags ($\gI$) & Number &  ~ \\ \hline
\textbf{Easy} & $\gI_a^{\text{(g)}}$, $\gI_a^{\text{(mpm)}}$ & 2 &$o^{\text{(dm)}}$  \\ 
\textbf{Standard} & $\gI_a^{\text{(g)}}$, $\gI_a^{\text{(s)}}$ &4 & $o^{\text{(dm)}}$\\ 
\textbf{Hard} & $\gI_a^{\text{(g)}}$, $\gI_a^{\text{(h)}}$ & 9 & $o^{\text{(dm)}}$ \\ 
\specialrule{0.8pt}{0pt}{2pt}
\end{tabular}
\caption{Statistics of various settings.}
\label{tab:task_complexity}
\end{table}
To assess MLLMs across varying levels of analytical complexity, we introduce a three-tiered evaluation framework.
The most challenging ``Hard'' setting employs raw remote-sensing data $\gI_a^{\text{(h)}}$, from GSWA without preprocessing steps. This configuration rigorously tests MLLMs' fundamental reasoning capabilities, demanding comprehensive interpretation with minimal prior knowledge.
In the ``Standard'' setting, MineBench preprocesses $\gI_a^{(\text{h})}$ into deposit signatures $\gI_a^{\text{(s)}}$, 
exposuring intuitive mineralization patterns.
The ``Easy'' setting, further simplify the ``Standard'' MineBench by using a manually preprocessed mineral prospectivity map  $\gI_a^{\text{(mpm)}}$, incorporating extensive prior geological knowledge.
MineBench statistics and task configurations are summarized in Table \ref{tab:task_complexity}.

\begin{figure}[thbp]
\centering
\includegraphics[width=0.8\linewidth]{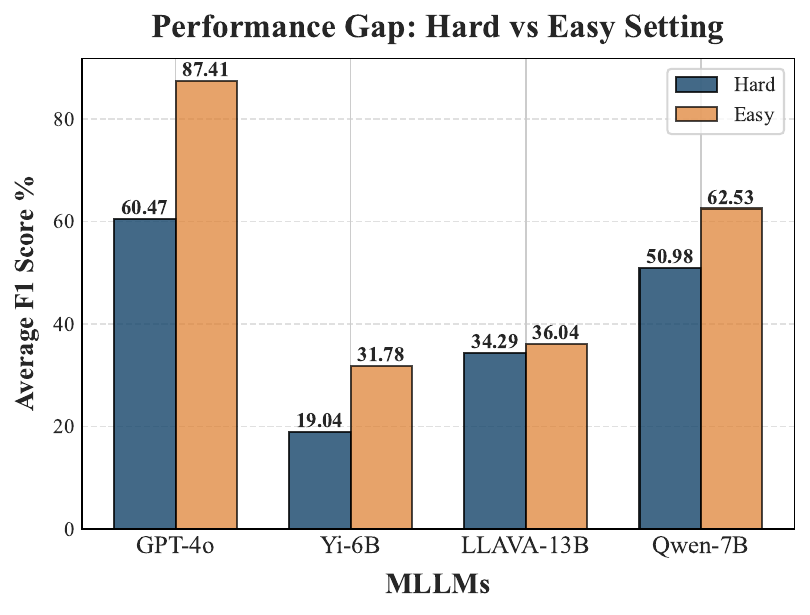}
    \caption{\small Performance across varying complexity levels}
    \label{fig:gap}
\end{figure}

Experimental results reveal a clear correlation between task complexity and MLLM performance, as shown in Figure \ref{fig:gap}. As settings demand more sophisticated domain expertise and multi-image reasoning, performance metrics systematically decline.
Even the state-of-the-art GPT-4o demonstrates this trend, with Avg.F1 scores declining from 87.41\% in the ``Easy'' task to 60.47\% in the ``Hard'' task. 
These findings underscore MineBench's critical role in identifying and facilitating improvements to current MLLM limitations by offering diverse, strategically designed evaluation settings.

\subsection{Alignment of MLLMs with Human}
\label{sec:consistence_eva}
\begin{figure}[t]
    \includegraphics[width=1\linewidth]{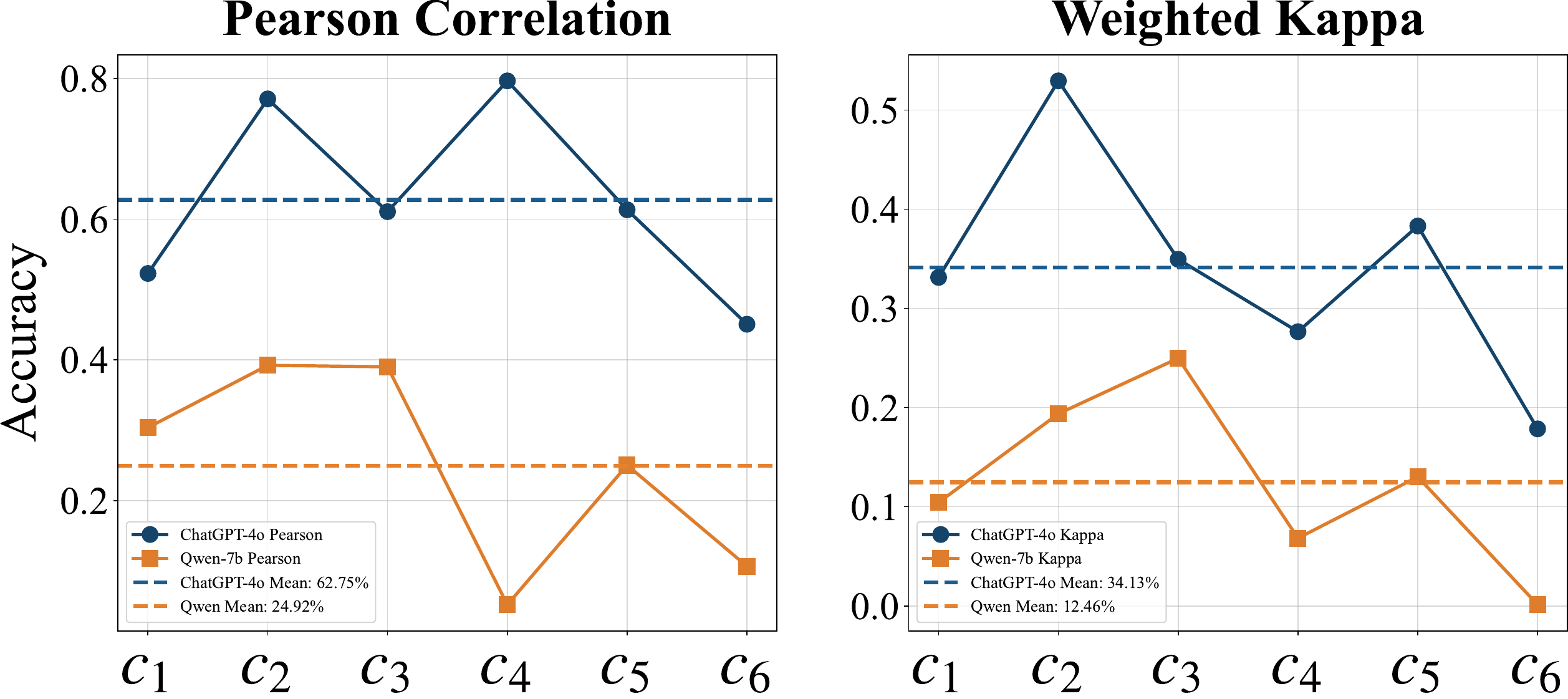}
    \caption{\small comparison of MLLMs and human assessment w.r.t Pearson correlation and weighted kappa across six outputs ($c_1$ to $c_6$). The dashed lines indicate avg. performance.}
    \label{fig:kappa}
\end{figure}

Following scoring guidelines, both the human expert and the MLLM are tasked with assigning a score within defined areas to evaluate reasoning ability in this evaluation.
The evaluation employed 20\% of the MineBench, with samples selected randomly while maintaining the original positive-negative class distribution. The quantitative results (Figure \ref{fig:kappa}) reveal two key findings:

\paragraph{Model-Human Agreement.} 
GPT-4o achieved significantly higher Pearson correlation and weighted Kappa scores than Qwen-7B across all evaluation criteria 
This strong alignment with human expert scoring patterns correlates with the models' overall performance, validating that higher-performing models better approximate human evaluation strategies~\cite{ouyang2022training}.

\paragraph{Challenges in Complex Task.} The output $c_6$ exhibited the lowest consistency scores, falling well below the average. 
This result underscores a key limitation: even advanced methods face difficulties extracting features from multi-image reasoning, which emphasizes the current shortcomings of MLLMs in handling complex mineral exploration scenarios. Additionally, the score distributions are visualized in Appendix~\ref{sec:score_distribution}, providing a comprehensive overview of their respective scoring tendencies. The consistency between scores and explanations in the MLLM responses is evaluated in Appendix~\ref{sec:score_explanation_consistency}.

\subsection{Ablation Studies}

\begin{table}[ht]
\centering
\small
\renewcommand{\arraystretch}{1.3} % Adjusts row height
\setlength{\tabcolsep}{2.4pt} % Adjusts column width
\begin{tabular}{c c c c c c }
\hline
Setting & Component   & Pos.F1 & Avg.F1 & Roc-Auc & Mcc \\ \hline
% \multirow{4}{*}{Hard}    & MineAgent  & \textbf{36.51}     & \underline{60.47}   & \textbf{68.63}   & \textbf{27.07} \\
%                         & w/o J   & 32.54     & 59.32   & 63.83   & 21.56 \\ 
%                         & w/o JC     & 25.53     & 49.08   & 58.92   & 11.72 \\ 
%                         & BASE       & 33.82     & 62.78   & 62.04   & 25.69 \\  \hline
\multirow{4}{*}{Standard}  & MineAgent  & \textbf{61.20}     & \textbf{77.19}   & \textbf{83.35}   & \textbf{56.30} \\
                        & w/o J  & 54.73     & 72.92   & 80.90   & 49.25 \\ 
                         & w/o JC     & 32.51     & 59.05   & 82.28   & 34.94 \\
                        & BASE       & 34.93     & 57.27   & 69.35   & 26.16 \\  \hline
\multirow{4}{*}{Easy}     & MineAgent  & 71.62     & 83.86   & 84.26   & 67.73 \\
                         & w/o A  & 44.00     & 62.91   & 80.27   & 40.62 \\ 
                        & w/o JC     & \textbf{77.78}     & \textbf{87.41}   & \textbf{86.96}   & \textbf{74.82} \\ 
                       & BASE       & 55.24     & 75.52   & 69.58   & 57.03 \\    \hline
\end{tabular}
\caption{\small Performance across different settings and components. The highest scores are marked in bold.}
\label{tab:ablation}
\end{table}

We conducted ablation studies to evaluate the effectiveness of MineAgent by analyzing its components using GPT-4o. Four variants were designed and evaluated to investigate the role of each component in handling tasks of different complexity levels. 
1) \emph{MineAgent:} The framework incorporates all components, including the judging modules, communication protocol, and decision-making modules. 
2) \emph{w/o J:} A variant of MineAgent that removes the judging modules, implementing the exploration pipeline in a single inference.
3) \emph{w/o JC:} Extending w/o J, this version further excludes the communication protocol, resulting in the absence of detailed scoring guidelines.(4) \emph{BASE:} A simplified version that additionally excludes the decision-making module, producing $o^{\text{(pipeline)}}$ as the result.

The results show MineAgent effectively reduces reasoning complexity by decomposing the pipeline into manageable components: The removal of judging modules led to a 7.05\% decline in the MCC score. Excluding the detailed scoring guidelines within the communication protocol caused a substantial performance drop.
% , with the MCC decreasing from 49.25\% to 34.94\%. 
Further, the decision-making modules played a critical role in enhancing the MLLM's capability to navigate the exploration pipeline. The details of the decision-making modules are further analyzed in Appendix \ref{sec:weight}, showing their effectiveness.
In the ``Easy'' setting, an unexpected performance pattern emerged: the \emph{w/o JC} variant outperformed the full MineAgent framework. This result aligns with prior findings: while existing MLLMs excel in basic visual reasoning, they often struggle with complex multi-image and domain-specific tasks~\cite{kazemi2024remi}.

\section{Conclusion}
In this paper, we present MineAgent, a novel MLLM agent framework designed to address critical challenges in multi-image reasoning and domain-specific gap for mineral exploration. 
% By integrating a local-global visual reasoning module and a collaborative evaluation mechanism, MineAgent offers a systematic approach to automating complex geospatial exploration tasks.
Our comprehensive quantitative and qualitative ablation studies further validate the effectiveness of MineAgent.
Further, our results underscore both the potential and the limitations of MLLMs in mineral exploration, revealing significant performance degradation as task complexity increases. 
% We hope this work serves as a foundation for advancing research in mineral exploration and inspires novel strategies to tackle geospatial reasoning and domain adaptation in MLLMs.

\section*{Limitation}
1) \emph{Generalized applications:} This work can only recognize specific types of deposits, restricting its applicability to a wider range of mineral types. 
2) \emph{Domain-Specific Knowledge Enhancement:} Continuing from the initial success of MineAgent, future work will explore strategies such as integrating domain-specific knowledge bases or leveraging reinforcement learning to further improve the MLLMs' performance in specialized tasks. 
3) \emph{Assistant Tools :} The results under the ``Hard'' setting highlight significant challenges. 
Future research will integrate assistant tools, such as integrating coding agents or feedback loops, can enhance the robustness and reasoning capabilities of models in realistic environments. 
% \subsection{Error Analysis}
% \textbf{Detail Misunderstanding:} This case study reveals a domain-specific limitation in the model's visual data interpretation. Figure \ref{fig:case}(a) depicts a mineralogical image with color-coded value regions.
% The machine-human assessment divergence is stark. The model assigned a score of 0, indicating no significant high-value regions, while the human evaluator scored 3. This discrepancy stems from the human's ability to interpret contextual nuances that computational analysis missed. 
% % Specifically, the expert recognized the importance of moderate-value areas (green) and low-value region distribution (blue)—key parameters in the specialized scoring framework.
% These overlooked elements are crucial within the specialized mineralogical scoring criteria, highlighting a significant gap in the model's domain-specific understanding.

% \textbf{Multi-image Reasoning:} Figure \ref{fig:case}(b) exposes another critical model limitation: compromised multi-image reasoning. Unlike the previous example of local detail misinterpretation, this case demonstrates the model's systemic difficulty in synthesizing spatial relationships across multiple images.
% In contrast, the human evaluator exhibited a more sophisticated evaluation.  By recognizing and integrating spatial relationships across different alteration indices, the expert revealed a holistic understanding that transcends the model's fragmented, image-by-image processing.

\bibliography{custom.bib}

\begin{thebibliography}{58}
\providecommand{\natexlab}[1]{#1}

\bibitem[{Alzubaidi et~al.(2021)Alzubaidi, Mostaghimi, Swietojanski, Clark, and Armstrong}]{cnn1}
Fatimah Alzubaidi, Peyman Mostaghimi, Pawel Swietojanski, Stuart~R Clark, and Ryan~T Armstrong. 2021.
\newblock Automated lithology classification from drill core images using convolutional neural networks.
\newblock \emph{Journal of Petroleum Science and Engineering}, 197:107933.

\bibitem[{Bai et~al.(2023)Bai, Bai, Yang, Wang, Tan, Wang, Lin, Zhou, and Zhou}]{Qwen-VL}
Jinze Bai, Shuai Bai, Shusheng Yang, Shijie Wang, Sinan Tan, Peng Wang, Junyang Lin, Chang Zhou, and Jingren Zhou. 2023.
\newblock Qwen-vl: A versatile vision-language model for understanding, localization, text reading, and beyond.
\newblock \emph{arXiv preprint arXiv:2308.12966}.

\bibitem[{Bedini(2017)}]{Bedini2017use}
Enton Bedini. 2017.
\newblock The use of hyperspectral remote sensing for mineral exploration: A review.
\newblock \emph{Journal of Hyperspectral Remote Sensing}, 7(4):189--211.

\bibitem[{Carranza(2008)}]{carranza2008geochemical}
Emmanuel John~Muico Carranza. 2008.
\newblock \emph{Geochemical anomaly and mineral prospectivity mapping in GIS}.
\newblock Elsevier.

\bibitem[{Chen et~al.(2024)Chen, Wu, Wang, Su, Chen, Xing, Zhong, Zhang, Zhu, Lu et~al.}]{chen2024internvl}
Zhe Chen, Jiannan Wu, Wenhai Wang, Weijie Su, Guo Chen, Sen Xing, Muyan Zhong, Qinglong Zhang, Xizhou Zhu, Lewei Lu, et~al. 2024.
\newblock Internvl: Scaling up vision foundation models and aligning for generic visual-linguistic tasks.
\newblock In \emph{Proceedings of the IEEE/CVF Conference on Computer Vision and Pattern Recognition}, pages 24185--24198.

\bibitem[{Cheng et~al.(2022)Cheng, Huang, Xu, Zhou, Li, and Wang}]{cheng2022nwpu}
Qimin Cheng, Haiyan Huang, Yuan Xu, Yuzhuo Zhou, Huanying Li, and Zhongyuan Wang. 2022.
\newblock Nwpu-captions dataset and mlca-net for remote sensing image captioning.
\newblock \emph{IEEE Transactions on Geoscience and Remote Sensing}, 60:1--19.

\bibitem[{Chicco and Jurman(2020)}]{mcc}
Davide Chicco and Giuseppe Jurman. 2020.
\newblock The advantages of the matthews correlation coefficient (mcc) over f1 score and accuracy in binary classification evaluation.
\newblock \emph{BMC genomics}, 21:1--13.

\bibitem[{Dentith and Mudge(2014)}]{dentith2024geophysics}
Michael Dentith and Stephen~T Mudge. 2014.
\newblock \emph{Geophysics for the mineral exploration geoscientist}.
\newblock Cambridge University Press.

\bibitem[{Du et~al.(2024)Du, Xiao, and Li}]{du2024haloscope}
Xuefeng Du, Chaowei Xiao, and Yixuan Li. 2024.
\newblock Haloscope: Harnessing unlabeled llm generations for hallucination detection.
\newblock \emph{arXiv preprint arXiv:2409.17504}.

\bibitem[{Fu et~al.(2023)Fu, Cheng, Jing, Ye, and Fu}]{cnn2}
Yufeng Fu, Qiuming Cheng, Linhai Jing, Bei Ye, and Hanze Fu. 2023.
\newblock Mineral prospectivity mapping of porphyry copper deposits based on remote sensing imagery and geochemical data in the duolong ore district, tibet.
\newblock \emph{Remote Sensing}, 15(2):439.

\bibitem[{Ghamisi et~al.(2017)Ghamisi, Yokoya, Li, Liao, Liu, Plaza, Rasti, and Plaza}]{ghamisi2017advances}
Pedram Ghamisi, Naoto Yokoya, Jun Li, Wenzhi Liao, Sicong Liu, Javier Plaza, Behnood Rasti, and Antonio Plaza. 2017.
\newblock Advances in hyperspectral image and signal processing: A comprehensive overview of the state of the art.
\newblock \emph{IEEE Geoscience and Remote Sensing Magazine}, 5(4):37--78.

\bibitem[{Gocht et~al.(2012)Gocht, Zantop, and Eggert}]{gocht2012international}
Werner~R Gocht, Half Zantop, and Roderick~G Eggert. 2012.
\newblock \emph{International mineral economics: mineral exploration, mine valuation, mineral markets, international mineral policies}.
\newblock Springer Science \& Business Media.

\bibitem[{Gonzalez-Alvarez et~al.(2020)Gonzalez-Alvarez, Goncalves, and Carranza}]{16}
Ignacio Gonzalez-Alvarez, MA~Goncalves, and Emmanuel John~M Carranza. 2020.
\newblock Introduction to the special issue challenges for mineral exploration in the 21st century: Targeting mineral deposits under cover.
\newblock \emph{Ore Geology Reviews}, 126:103785.

\bibitem[{Hewson et~al.(2015)Hewson, Robson, Mauger, Cudahy, Thomas, and Jones}]{15}
Robert Hewson, D~Robson, Alan Mauger, Thomas Cudahy, Matilda Thomas, and Simon Jones. 2015.
\newblock Using the geoscience australia-csiro aster maps and airborne geophysics to explore australian geoscience.
\newblock \emph{Journal of Spatial Science}, 60(2):207--231.

\bibitem[{Jiang et~al.(2024)Jiang, He, Zeng, Wei, Ku, Liu, and Chen}]{jiang2024mantis}
Dongfu Jiang, Xuan He, Huaye Zeng, Cong Wei, Max Ku, Qian Liu, and Wenhu Chen. 2024.
\newblock Mantis: Interleaved multi-image instruction tuning.
\newblock \emph{arXiv preprint arXiv:2405.01483}.

\bibitem[{Kazemi et~al.(2024)Kazemi, Dikkala, Anand, Devic, Dasgupta, Liu, Fatemi, Awasthi, Guo, Gollapudi et~al.}]{kazemi2024remi}
Mehran Kazemi, Nishanth Dikkala, Ankit Anand, Petar Devic, Ishita Dasgupta, Fangyu Liu, Bahare Fatemi, Pranjal Awasthi, Dee Guo, Sreenivas Gollapudi, et~al. 2024.
\newblock Remi: A dataset for reasoning with multiple images.
\newblock \emph{arXiv preprint arXiv:2406.09175}.

\bibitem[{Kuckreja et~al.(2024)Kuckreja, Danish, Naseer, Das, Khan, and Khan}]{kuckreja2024geochat}
Kartik Kuckreja, Muhammad~Sohail Danish, Muzammal Naseer, Abhijit Das, Salman Khan, and Fahad~Shahbaz Khan. 2024.
\newblock Geochat: Grounded large vision-language model for remote sensing.
\newblock In \emph{Proceedings of the IEEE/CVF Conference on Computer Vision and Pattern Recognition}, pages 27831--27840.

\bibitem[{Li et~al.(2024)Li, Xu, Mei, Hua, Rama, Raheja, Wang, Zhu, and Zhang}]{li2024autoflow}
Zelong Li, Shuyuan Xu, Kai Mei, Wenyue Hua, Balaji Rama, Om~Raheja, Hao Wang, He~Zhu, and Yongfeng Zhang. 2024.
\newblock Autoflow: Automated workflow generation for large language model agents.
\newblock \emph{arXiv preprint arXiv:2407.12821}.

\bibitem[{Liu et~al.(2023{\natexlab{a}})Liu, Wang, Tang, Wang, Zheng, Sun, Zhang, Gan, and Cao}]{liu2023deep}
Cai Liu, Wenlei Wang, Juxing Tang, Qin Wang, Ke~Zheng, Yanyun Sun, Jiahong Zhang, Fuping Gan, and Baobao Cao. 2023{\natexlab{a}}.
\newblock A deep-learning-based mineral prospectivity modeling framework and workflow in prediction of porphyry--epithermal mineralization in the duolong ore district, tibet.
\newblock \emph{Ore Geology Reviews}, 157:105419.

\bibitem[{Liu et~al.(2024{\natexlab{a}})Liu, Li, Li, Li, Zhang, Shen, and Lee}]{liu2024llavanext}
Haotian Liu, Chunyuan Li, Yuheng Li, Bo~Li, Yuanhan Zhang, Sheng Shen, and Yong~Jae Lee. 2024{\natexlab{a}}.
\newblock \href {https://llava-vl.github.io/blog/2024-01-30-llava-next/} {Llava-next: Improved reasoning, ocr, and world knowledge}.

\bibitem[{Liu et~al.(2023{\natexlab{b}})Liu, Li, Wu, and Lee}]{liu2023llava}
Haotian Liu, Chunyuan Li, Qingyang Wu, and Yong~Jae Lee. 2023{\natexlab{b}}.
\newblock Visual instruction tuning.

\bibitem[{Liu et~al.(2024{\natexlab{b}})Liu, Zhang, Xu, Shi, Jiang, Yan, Zhang, Huang, Yuan, Li et~al.}]{liu2024mibench}
Haowei Liu, Xi~Zhang, Haiyang Xu, Yaya Shi, Chaoya Jiang, Ming Yan, Ji~Zhang, Fei Huang, Chunfeng Yuan, Bing Li, et~al. 2024{\natexlab{b}}.
\newblock Mibench: Evaluating multimodal large language models over multiple images.
\newblock \emph{arXiv preprint arXiv:2407.15272}.

\bibitem[{Liu et~al.(2024{\natexlab{c}})Liu, Lin, Hewitt, Paranjape, Bevilacqua, Petroni, and Liang}]{liu2024lost}
Nelson~F Liu, Kevin Lin, John Hewitt, Ashwin Paranjape, Michele Bevilacqua, Fabio Petroni, and Percy Liang. 2024{\natexlab{c}}.
\newblock Lost in the middle: How language models use long contexts.
\newblock \emph{Transactions of the Association for Computational Linguistics}, 12:157--173.

\bibitem[{Liu et~al.(2024{\natexlab{d}})Liu, Chu, Zang, Wei, Dong, Zhang, Liang, Xiong, Qiao, Lin et~al.}]{13}
Ziyu Liu, Tao Chu, Yuhang Zang, Xilin Wei, Xiaoyi Dong, Pan Zhang, Zijian Liang, Yuanjun Xiong, Yu~Qiao, Dahua Lin, et~al. 2024{\natexlab{d}}.
\newblock Mmdu: A multi-turn multi-image dialog understanding benchmark and instruction-tuning dataset for lvlms.
\newblock \emph{arXiv preprint arXiv:2406.11833}.

\bibitem[{Luo et~al.(2024)Luo, Pang, Zhang, Wang, Wang, Dang, Lao, Wang, Chen, Tan et~al.}]{luo2024skysensegpt}
Junwei Luo, Zhen Pang, Yongjun Zhang, Tingzhu Wang, Linlin Wang, Bo~Dang, Jiangwei Lao, Jian Wang, Jingdong Chen, Yihua Tan, et~al. 2024.
\newblock Skysensegpt: A fine-grained instruction tuning dataset and model for remote sensing vision-language understanding.
\newblock \emph{arXiv preprint arXiv:2406.10100}.

\bibitem[{Muhtar et~al.(2024)Muhtar, Li, Gu, Zhang, and Xiao}]{10}
Dilxat Muhtar, Zhenshi Li, Feng Gu, Xueliang Zhang, and Pengfeng Xiao. 2024.
\newblock Lhrs-bot: Empowering remote sensing with vgi-enhanced large multimodal language model.
\newblock \emph{arXiv preprint arXiv:2402.02544}.

\bibitem[{OpenAI(2024)}]{chatgpt4o}
OpenAI. 2024.
\newblock Chatgpt-4o.
\newblock Available at \url{https://openai.com}.

\bibitem[{Ousmanou et~al.(2024)Ousmanou, Martial, Jules, Ludovic, Blandine, Sufinatu, Mohamed, and Maurice}]{Ousmanou2024Mapping}
Safianou Ousmanou, Fozing~Eric Martial, Tcheumenak~Kou{\'e}mo Jules, Achu~Megnemo Ludovic, Kamgang Tchuifong~Agn{\`e}s Blandine, Aman Sufinatu, Rachid Mohamed, and Kw{\'e}kam Maurice. 2024.
\newblock Mapping and discrimination of the mineralization potential in granitoids from banyo area (adamawa, cameroon), using landsat 9 oli, aster images and field observations.
\newblock \emph{Geosystems and Geoenvironment}, 3(1):100239.

\bibitem[{Ouyang et~al.(2022)Ouyang, Wu, Jiang, Almeida, Wainwright, Mishkin, Zhang, Agarwal, Slama, Ray et~al.}]{ouyang2022training}
Long Ouyang, Jeffrey Wu, Xu~Jiang, Diogo Almeida, Carroll Wainwright, Pamela Mishkin, Chong Zhang, Sandhini Agarwal, Katarina Slama, Alex Ray, et~al. 2022.
\newblock Training language models to follow instructions with human feedback.
\newblock \emph{Advances in neural information processing systems}, 35:27730--27744.

\bibitem[{Pham et~al.(2018)Pham, Prakash, and Bui}]{pham2018spatial}
Binh~Thai Pham, Indra Prakash, and Dieu~Tien Bui. 2018.
\newblock Spatial prediction of landslides using a hybrid machine learning approach based on random subspace and classification and regression trees.
\newblock \emph{Geomorphology}, 303:256--270.

\bibitem[{Portal()}]{dataset}
AuScope~Discovery Portal.
\newblock Satellite aster geoscience product notes for australia.

\bibitem[{Ren et~al.(2020)Ren, Sun, and Zhai}]{ren2020improved}
Zhongliang Ren, Lin Sun, and Qiuping Zhai. 2020.
\newblock Improved k-means and spectral matching for hyperspectral mineral mapping.
\newblock \emph{International Journal of Applied Earth Observation and Geoinformation}, 91:102154.

\bibitem[{Sabins(1999)}]{Sabins1999Remote}
Floyd~F Sabins. 1999.
\newblock Remote sensing for mineral exploration.
\newblock \emph{Ore geology reviews}, 14(3-4):157--183.

\bibitem[{Shirmard et~al.(2022)Shirmard, Farahbakhsh, M{\"u}ller, and Chandra}]{mineral_review}
Hojat Shirmard, Ehsan Farahbakhsh, R~Dietmar M{\"u}ller, and Rohitash Chandra. 2022.
\newblock A review of machine learning in processing remote sensing data for mineral exploration.
\newblock \emph{Remote Sensing of Environment}, 268:112750.

\bibitem[{Snoek et~al.(2012)Snoek, Larochelle, and Adams}]{bayesian}
Jasper Snoek, Hugo Larochelle, and Ryan~P Adams. 2012.
\newblock Practical bayesian optimization of machine learning algorithms.
\newblock \emph{Advances in neural information processing systems}, 25.

\bibitem[{Team et~al.(2024)Team, Georgiev, Lei, Burnell, Bai, Gulati, Tanzer, Vincent, Pan, Wang et~al.}]{team2024gemini}
Gemini Team, Petko Georgiev, Ving~Ian Lei, Ryan Burnell, Libin Bai, Anmol Gulati, Garrett Tanzer, Damien Vincent, Zhufeng Pan, Shibo Wang, et~al. 2024.
\newblock Gemini 1.5: Unlocking multimodal understanding across millions of tokens of context.
\newblock \emph{arXiv preprint arXiv:2403.05530}.

\bibitem[{Van~der Maaten and Hinton(2008)}]{van2008visualizing}
Laurens Van~der Maaten and Geoffrey Hinton. 2008.
\newblock Visualizing data using t-sne.
\newblock \emph{Journal of machine learning research}, 9(11).

\bibitem[{Van~der Meer et~al.(2012)Van~der Meer, Van~der Werff, Van~Ruitenbeek, Hecker, Bakker, Noomen, Van Der~Meijde, Carranza, De~Smeth, and Woldai}]{Van2012Multi}
Freek~D Van~der Meer, Harald~MA Van~der Werff, Frank~JA Van~Ruitenbeek, Chris~A Hecker, Wim~H Bakker, Marleen~F Noomen, Mark Van Der~Meijde, E~John~M Carranza, J~Boudewijn De~Smeth, and Tsehaie Woldai. 2012.
\newblock Multi-and hyperspectral geologic remote sensing: A review.
\newblock \emph{International journal of applied Earth observation and geoinformation}, 14(1):112--128.

\bibitem[{Wambo et~al.(2020)Wambo, Pour, Ganno, Asimow, Zoheir, dos Reis~Salles, Nzenti, Pradhan, and Muslim}]{wambo2020identifying}
Jonas Didero~Takodjou Wambo, Amin~Beiranvand Pour, Sylvestre Ganno, Paul~D Asimow, Basem Zoheir, Rodrigo dos Reis~Salles, Jean~Paul Nzenti, Biswajeet Pradhan, and Aidy~M Muslim. 2020.
\newblock Identifying high potential zones of gold mineralization in a sub-tropical region using landsat-8 and aster remote sensing data: a case study of the ngoura-colomines goldfield, eastern cameroon.
\newblock \emph{Ore Geology Reviews}, 122:103530.

\bibitem[{Wang et~al.(2024{\natexlab{a}})Wang, Zhang, Du, Xu, Liu, Tao, and Zhang}]{12}
Di~Wang, Jing Zhang, Bo~Du, Minqiang Xu, Lin Liu, Dacheng Tao, and Liangpei Zhang. 2024{\natexlab{a}}.
\newblock Samrs: Scaling-up remote sensing segmentation dataset with segment anything model.
\newblock \emph{Advances in Neural Information Processing Systems}, 36.

\bibitem[{Wang et~al.(2024{\natexlab{b}})Wang, Fu, Huang, Li, Liu, Liu, Ma, Xu, Zhou, Zhang et~al.}]{wang2024muirbench}
Fei Wang, Xingyu Fu, James~Y Huang, Zekun Li, Qin Liu, Xiaogeng Liu, Mingyu~Derek Ma, Nan Xu, Wenxuan Zhou, Kai Zhang, et~al. 2024{\natexlab{b}}.
\newblock Muirbench: A comprehensive benchmark for robust multi-image understanding.
\newblock \emph{arXiv preprint arXiv:2406.09411}.

\bibitem[{Wang et~al.(2021)Wang, Zheng, Ma, Lu, and Zhong}]{wang2021loveda}
Junjue Wang, Zhuo Zheng, Ailong Ma, Xiaoyan Lu, and Yanfei Zhong. 2021.
\newblock Loveda: A remote sensing land-cover dataset for domain adaptive semantic segmentation.
\newblock \emph{arXiv preprint arXiv:2110.08733}.

\bibitem[{Wang et~al.(2024{\natexlab{c}})Wang, Bai, Tan, Wang, Fan, Bai, Chen, Liu, Wang, Ge, Fan, Dang, Du, Ren, Men, Liu, Zhou, Zhou, and Lin}]{Qwen2VL}
Peng Wang, Shuai Bai, Sinan Tan, Shijie Wang, Zhihao Fan, Jinze Bai, Keqin Chen, Xuejing Liu, Jialin Wang, Wenbin Ge, Yang Fan, Kai Dang, Mengfei Du, Xuancheng Ren, Rui Men, Dayiheng Liu, Chang Zhou, Jingren Zhou, and Junyang Lin. 2024{\natexlab{c}}.
\newblock Qwen2-vl: Enhancing vision-language model's perception of the world at any resolution.
\newblock \emph{arXiv preprint arXiv:2409.12191}.

\bibitem[{Wang et~al.(2024{\natexlab{d}})Wang, Zhou, Liu, Lu, Xu, He, Yoon, Lu, Bertasius, Bansal et~al.}]{14}
Xiyao Wang, Yuhang Zhou, Xiaoyu Liu, Hongjin Lu, Yuancheng Xu, Feihong He, Jaehong Yoon, Taixi Lu, Gedas Bertasius, Mohit Bansal, et~al. 2024{\natexlab{d}}.
\newblock Mementos: A comprehensive benchmark for multimodal large language model reasoning over image sequences.
\newblock \emph{arXiv preprint arXiv:2401.10529}.

\bibitem[{Xu et~al.(2021)Xu, Li, Xie, Cai, Niu, and Liu}]{xu2021mineral}
Yongyang Xu, Zixuan Li, Zhong Xie, Huihui Cai, Pengfei Niu, and Hui Liu. 2021.
\newblock Mineral prospectivity mapping by deep learning method in yawan-daqiao area, gansu.
\newblock \emph{Ore Geology Reviews}, 138:104316.

\bibitem[{Yang et~al.(2024)Yang, Zuo, and Kreuzer}]{8}
Fanfan Yang, Renguang Zuo, and Oliver~P Kreuzer. 2024.
\newblock Artificial intelligence for mineral exploration: A review and perspectives on future directions from data science.
\newblock \emph{Earth-Science Reviews}, page 104941.

\bibitem[{Yazdi et~al.(2018)Yazdi, Jafari~Rad, Aghazadeh, and Afzal}]{alteration}
Zahra Yazdi, Alireza Jafari~Rad, Mehraj Aghazadeh, and Peyman Afzal. 2018.
\newblock Alteration mapping for porphyry copper exploration using aster and quickbird multispectral images, sonajeel prospect, nw iran.
\newblock \emph{Journal of the Indian Society of Remote Sensing}, 46:1581--1593.

\bibitem[{Yin et~al.(2024{\natexlab{a}})Yin, Long, Liu, Khalil, and Ye}]{transformer1}
Chuntao Yin, Yaqian Long, Lei Liu, Yasir~Shaheen Khalil, and Songxing Ye. 2024{\natexlab{a}}.
\newblock Mapping ni-cu-platinum group element-hosting, small-sized, mafic-ultramafic rocks using worldview-3 images and a spatial-spectral transformer deep learning method.
\newblock \emph{Economic Geology}, 119(3):665--680.

\bibitem[{Yin et~al.(2024{\natexlab{b}})Yin, Fu, Zhao, Li, Sun, Xu, and Chen}]{9}
Shukang Yin, Chaoyou Fu, Sirui Zhao, Ke~Li, Xing Sun, Tong Xu, and Enhong Chen. 2024{\natexlab{b}}.
\newblock A survey on multimodal large language models.
\newblock \emph{National Science Review}, page nwae403.

\bibitem[{Young et~al.(2024)Young, Chen, Li, Huang, Zhang, Zhang, Li, Zhu, Chen, Chang et~al.}]{yi}
Alex Young, Bei Chen, Chao Li, Chengen Huang, Ge~Zhang, Guanwei Zhang, Heng Li, Jiangcheng Zhu, Jianqun Chen, Jing Chang, et~al. 2024.
\newblock Yi: Open foundation models by 01. ai.
\newblock \emph{arXiv preprint arXiv:2403.04652}.

\bibitem[{Zhan et~al.(2023)Zhan, Xiong, and Yuan}]{zhan2023rsvg}
Yang Zhan, Zhitong Xiong, and Yuan Yuan. 2023.
\newblock Rsvg: Exploring data and models for visual grounding on remote sensing data.
\newblock \emph{IEEE Transactions on Geoscience and Remote Sensing}, 61:1--13.

\bibitem[{Zhan et~al.(2024)Zhan, Xiong, and Yuan}]{zhan2024skyeyegpt}
Yang Zhan, Zhitong Xiong, and Yuan Yuan. 2024.
\newblock Skyeyegpt: Unifying remote sensing vision-language tasks via instruction tuning with large language model.
\newblock \emph{arXiv preprint arXiv:2401.09712}.

\bibitem[{Zhang et~al.(2024{\natexlab{a}})Zhang, Yang, Lyu, Jin, Yao, Chen, and Luo}]{zhang2024cocot}
Daoan Zhang, Junming Yang, Hanjia Lyu, Zijian Jin, Yuan Yao, Mingkai Chen, and Jiebo Luo. 2024{\natexlab{a}}.
\newblock Cocot: Contrastive chain-of-thought prompting for large multimodal models with multiple image inputs.
\newblock \emph{arXiv preprint arXiv:2401.02582}.

\bibitem[{Zhang et~al.(2024{\natexlab{b}})Zhang, Yu, Dong, Li, Su, Chu, and Yu}]{zhang2024mm}
Duzhen Zhang, Yahan Yu, Jiahua Dong, Chenxing Li, Dan Su, Chenhui Chu, and Dong Yu. 2024{\natexlab{b}}.
\newblock Mm-llms: Recent advances in multimodal large language models.
\newblock \emph{arXiv preprint arXiv:2401.13601}.

\bibitem[{Zhang et~al.(2024{\natexlab{c}})Zhang, Cai, Zhang, Zhuang, and Mao}]{11}
Wei Zhang, Miaoxin Cai, Tong Zhang, Yin Zhuang, and Xuerui Mao. 2024{\natexlab{c}}.
\newblock Earthgpt: A universal multi-modal large language model for multi-sensor image comprehension in remote sensing domain.
\newblock \emph{IEEE Transactions on Geoscience and Remote Sensing}.

\bibitem[{Zhao et~al.(2024)Zhao, Zong, Zhang, and Hospedales}]{zhao2024benchmarking}
Bingchen Zhao, Yongshuo Zong, Letian Zhang, and Timothy Hospedales. 2024.
\newblock Benchmarking multi-image understanding in vision and language models: Perception, knowledge, reasoning, and multi-hop reasoning.
\newblock \emph{arXiv preprint arXiv:2406.12742}.

\bibitem[{Zuo(2020)}]{zuo2020geodata}
Renguang Zuo. 2020.
\newblock Geodata science-based mineral prospectivity mapping: A review.
\newblock \emph{Natural Resources Research}, 29(6):3415--3424.

\bibitem[{Zuo et~al.(2021)Zuo, Kreuzer, Wang, Xiong, Zhang, and Wang}]{mpm}
Renguang Zuo, Oliver~P Kreuzer, Jian Wang, Yihui Xiong, Zhenjie Zhang, and Ziye Wang. 2021.
\newblock Uncertainties in gis-based mineral prospectivity mapping: Key types, potential impacts and possible solutions.
\newblock \emph{Natural Resources Research}, 30:3059--3079.

\end{thebibliography}
\newpage

\appendix

\section{Benchmark Construction}
\label{sec:construction}
\begin{table*}[ht!]
\centering
\scriptsize
\renewcommand{\arraystretch}{1.5}
\setlength{\tabcolsep}{7pt}
\begin{tabular}{l l m{7cm}l}
\hline
\textbf{Remote-sening Data} & \textbf{Associated Minerals} & \textbf{Geoscience Application} & \textbf{Notion}
\\ \hline
\textbf{Ferric Oxide Content} & Hematite, Goethite & Identifies oxidation zones within hydrothermal systems, where hematite and goethite accumulate due to surface weathering and high-oxidation conditions. Weakly associated with propylitic zones but commonly found in silicified zones within epithermal environments. &  $\gI_a^{\text{(h,ox)}}$
\\
\textbf{FeOH Group Content} & Jarosite, Chlorite, Epidote & Indicates FeOH-bearing minerals typical of acid-sulfate environments in hydrothermal systems. Strongly associated with propylitic alteration, with chlorite-epidote assemblages marking the transition between hydrothermal and propylitic zones. &  $\gI_a^{\text{(h,oh)}}$ 
\\ 
\textbf{Opaque Index} & Magnetite, Pyrite, Manganese Oxides & Highlights reduced zones containing opaque minerals like magnetite and pyrite. Primarily found in the core areas of hydrothermal systems and occasionally in overprinted propylitic zones. &  $\gI_a^{\text{(h,op)}}$ 
\\ 
\textbf{AlOH Group Content } & Muscovite, Kaolinite, Montmorillonite & Identifies AlOH-rich clays commonly associated with phyllic alteration in hydrothermal systems and transitional zones between phyllic and propylitic alteration. & $\gI_a^{\text{(h,al)}}$ 
\\ 
\textbf{MgOH Group Content} & Chlorite, Epidote, Calcite & Detects MgOH-bearing minerals, which form broad halos around hydrothermal zones as part of propylitic alteration. Typically shows an inverse correlation with silicification and is essential for mapping zonal alteration patterns. & $\gI_a^{\text{(h,mg)}}$
\\ 
\textbf{Ferrous Iron Content} & Chlorite & Primarily identifies iron-rich minerals within potassic zones of hydrothermal systems, including biotite and magnetite. Common in hydrothermal cores and useful for distinguishing primary iron minerals from secondary phases. & $\gI_a^{\text{(h,fe)}}$
\\ 
\textbf{Quartz Index} & Quartz & Strongly correlated with silicification, particularly in quartz-dominant zones and silica-rich veins. Helps distinguish crystalline quartz from other forms of silica or silicates, such as feldspar, which is essential for mapping silicified alteration zones. & $\gI_a^{\text{(h,qa)}}$
\\ 
\textbf{Silica Index } & SiO$_2$, Quartz & A key indicator of silicification, especially in advanced argillic zones of hydrothermal systems. Common in quartzite, silicified cap rocks, and vein systems. &  $\gI_a^{\text{(h,si)}}$
\\ \hline
\textbf{False Color Image} & Geological Environments & Used to differentiate geological features from non-geological elements, such as vegetation, clouds, and shadows. Also serves as a greyscale or color background to visualize and interpret index-based alteration patterns. & $\gI_a^{\text{(g)}}$
\\ \hline
\end{tabular}
\caption{The details of nine remote-sensing data.}
\label{tab:raw_data}
\end{table*}

MineBench is based on the publicly available GSWA remote-sensing dataset, a reliable resource for geoscience applications. GSWA comprises 17 ASTER remote-sensing data, 14 of which are synthesized using ASTER's nine visible, near-infrared, and shortwave infrared bands (bands 1–9). The remaining three datasets utilize ASTER's thermal infrared bands (bands 10–14), extending the spectral range and enhancing suitability for geoscience interpretation. Each data underwent rigorous processing and evaluation to serve as the foundational base for MineBench.

\paragraph{Data Collection} 

To identify the targeted deposits, we selected nine remote-sensing datasets $\gI^a$ for analysis. The selected datasets $\gI^a$ are categorized into two subsets based on their roles: geological images $\gI_a^{\text{(g)}}$ and hyperspectral images $\gI_a^{\text{(h)}}$, as shown in Table \ref{tab:raw_data}. Details of these datasets are provided in the original technical document~\citet{dataset}. The remote-sensing data were segmented into 12$\times$12 $km^2$ areas using a grid method to facilitate detection. 
To ensure high-quality data, areas containing blurry, irrelevant images or those heavily affected by shadows obscuring geological features were excluded. After labeling, areas with mineral deposits near edges or other factors hindering accurate identification were excluded.

\paragraph{Quality Control with Human Annotators.}
A two-stage human review process was implemented to ensure data quality. A general reviewer conducted an initial quality check, followed by expert review and refinement. Areas requiring additional geochemical data or field observations for accurate identification were removed to improve prediction accuracy. This rigorous process ensures the accuracy and utility of MineBench.

\section{Data Preprocessing Workflow}
\label{sec:preprocess}

\begin{table*}[ht!]
\centering
\scriptsize
\renewcommand{\arraystretch}{1.6}
\setlength{\tabcolsep}{5.5pt}
\begin{tabular}{c c c c c c}
\hline
\textbf{MPM} & \textbf{\makecell{Deposit Signatures}} & \textbf{Raw Images} & \textbf{Value Range of Raw Images} & 
\textbf{Weight of Raw Images}  & \textbf{Weight of Signatures} \\ 
\hline
\multirow{10}{*}{\makecell{\textbf{Copper Deposit} \\ \\ \textbf{(Value Range: 0-5)}}} & Geological Environment & False Color Image &\multirow{1}{*}{0-1} & 1 &  1 \\ 
\cline{2-6}
& \multirow{3}{*}{\makecell{Hydrothermal Alteration \\ (Value Range: 1-3)}} & Ferric Oxide Content & 1.1 -- 2.1 & 1 & \multirow{3}{*}{5} \\ 
& & FeOH Group Content& 2.03 -- 2.25 & 2 & \\ 
& & Opaque Index & 0.4 -- 0.9 & 4 & \\ 
\cline{2-6}
& \multirow{4}{*}{\makecell{Propylitic Alteration \\ (Value Range: 0.6-1)}} & AlOH Group Content & 2.0 -- 2.25 & 1  & \multirow{4}{*} 3 \\ 
& & FeOH Group Content & 2.03 -- 2.25 & 1 & \\ 
& & MgOH Group Content & 1.05 -- 1.2 & 1 & \\ 
& & Ferrous Iron Content & 0.1 -- 2.0 & 2 & \\ 
\cline{2-6}
& \multirow{3}{*}{\makecell{Silicification Zone \\ (Value Range: 1-3)}} & Ferric Oxide Content & 1.1 -- 2.1 & 1 & \multirow{3}{*}{1} \\ 
& & Quartz Index &  1.0 -- 1.35 & 1 & \\ 
& & Silica Index & 0.5 -- 0.52 & 2 & \\ 
\hline
\end{tabular}
\caption{Weights and value ranges of deposit signatures and MPM in MineBench.}
\label{tab:weight}
\end{table*}

A preprocessing workflow is proposed to construct the deposit signatures and mineral prospectivity map (MPM) used in MineBench evaluation and validation.
This workflow enables precise control over feature extraction and integration, specifically optimized for copper mineralization detection. The detailed steps are as follows:

\subsection{Normalization of Remote-sensing Images}

To ensure consistency and standardization, each remote-sensing image $\gI_a$ undergoes min-max normalization:
\begin{equation}
    \gI_{\text{a,norm}} = \frac{\gI_a - \min(\gI_a)}{\max(\gI_a) - \min(\gI_a)}
\end{equation}
where $\min(\gI_a)$ and $\max(\gI_a)$ are obtained from the original dataset (Table~\ref{tab:weight}). This normalization scales remote-sensing images to the $[0,1]$ range, facilitating seamless integration in preprocessing steps.
Additionally, the normalization function $\text{norm}_{[a,b]}$ ensures a consistent scaling range throughout the workflow:
\begin{equation}
\text{norm}_{[a,b]}(\gI_a) = \begin{cases} 
0 & \text{if } x < a \\
\frac{\gI_a-a}{b-a} & \text{if } a \leq \gI_a \leq b \\
1 & \text{if } x > b
\end{cases}
\end{equation}
The function maps values to $[0,1]$, capping outliers beyond specified bounds while preserving linear scaling within the target range. It improves the identification of high-potential areas by ensuring standardized and consistent data scaling.

\subsection{Preprocessing Step}

\paragraph{Deposit Signatures.} After normalization, raw hyperspectral images $\gI_a^{\text{(h)}}$ are transformed into deposit signatures $\gI_a^{\text{(s)}}$ using a weighted linear combination informed by domain-specific knowledge (Table~\ref{tab:weight}). These relationships are critical for identifying regions with varying mineralization patterns.
The first signature, the hydrothermal alteration zone $\gI_a^{\text{(s,h)}}$, a primary indicator of copper mineralization, is computed as:
\begin{equation}
\centering
\gI_a^{\text{(s,h)}} =  \text{norm}_{[1,3]}(\gI_a^{\text{(h,ox)}} + 2\gI_a^{\text{(h,oh)}} + 4\gI_a^{\text{(h,op)}})
\end{equation}
where $\gI_a^{\text{(h,ox)}}$ represents ferric oxide content, $\gI_a^{\text{(h,oh)}}$ denotes FeOH group content, and $\gI_a^{\text{(h,op)}}$ indicates the opaque index. $\gI_a^{\text{(h,ox)}}$, $\gI_a^{\text{(h,oh)}}$, and $\gI_a^{\text{(h,op)}}$ are types of hyperspectral images associated with specific minerals. The weights 1, 2, and 4 reflect each indicator's relative significance in identifying hydrothermal alteration. The normalization function scales the output to the range $[1,3]$, enabling cross-regional comparisons and threshold-based targeting. The value range is derived from expert observations.

Similarly, the second signature, the propylitic alteration zone $\gI_a^{\text{(s,p)}}$, which characterizes peripheral mineralization areas, is calculated as:
\begin{equation}
\centering
\gI_a^{\text{(s,p)}} = \text{norm}_{[0.6,1]}(\gI_a^{\text{(h,al)}} + \gI_a^{\text{(h,oh)}} + \gI_a^{\text{(h,mg)}} + 2\gI_a^{\text{(h,fe)}})
\end{equation}
where $\gI_a^{\text{(h,al)}}$ represents AlOH group content, $\gI_a^{\text{(h,mg)}}$ indicates MgOH group content, and $\gI_a^{\text{(h,fe)}}$ denotes ferrous iron content.

The final signature, the silicification zone $\gI_a^{\text{(s,s)}}$, which indicates secondary mineralization patterns, is quantified as:
\begin{equation}
\centering
\gI_a^{\text{(s,s)}} = \text{norm}_{[1,2.5]}(\gI_a^{\text{(h,ox)}} + \gI_a^{\text{(h,qa)}} + 2\gI_a^{\text{(h,si)}})
\end{equation}
where $\gI_a^{\text{(h,qa)}}$ represents quartz content and $\gI_a^{\text{(h,si)}}$ indicates silica abundance. 
This weighted combination captures distinct deposit signatures essential for deposit detection.

\paragraph{Mineral Prospectivity Map.}
Following the weighted linear combination, the mineral potential map is constructed by combining deposit signatures $\gI_a^{\text{(s,h)}}$, $\gI_a^{\text{(s,p)}}$, and $\gI_a^{\text{(s,s)}}$ to quantitatively evaluate the copper deposit potential based on the spatial distribution and intensity of key alteration zones:
\begin{equation}
\gI_a^{\text{(mpm)}} = \text{normalize}_{[0,5]} (5\gI_a^{\text{(s,h)}} + 3\gI_a^{\text{(s,p)}} + \gI_a^{\text{(s,s)}})
\end{equation}
The weights (5, 3, and 1) are derived from extensive statistical analysis of known copper deposits across diverse geological settings, reflecting the relative contribution of each zone to copper mineralization.

\emph{Hydrothermal Alteration Zone.} The core zone $\gI_a^{\text{(s,h)}}$ is characterized by intense hydrothermal alteration and high-temperature mineral assemblages, exhibiting the strongest spatial correlation with copper mineralization.

\emph{Propylitic Alteration Zone.} Surrounding the hydrothermal core, the intermediate zone $\gI_a^{\text{(s,p)}}$ is characterized by moderate-temperature alteration minerals, including chlorite, epidote, and calcite. Although not directly mineralized, this zone provides critical context for delineating the extent of the hydrothermal system.

\emph{Silicification Zone.} The outermost zone $\gI_a^{\text{(s,s)}}$ is marked by silica enrichment and the presence of low-temperature minerals. While less directly associated with mineralization, this zone delineates system boundaries and fluid flow patterns.

This integrated approach highlights the hierarchical significance of different alteration signatures, improving the accuracy of copper mineral potential assessments.

\subsection{Data Visualization}
Finally, we get three types of mineralogical data: raw hyperspectral image $\gI_a^{\text{(h)}}$, deposit signatures $\gI_a^{\text{(s)}}$, and mineral prospectivity maps $\gI_a^{\text{(mpm)}}$.
To enhance the interpretability of these data, we employ a visualization process comprising several key steps. First, we implement transparency for data points with zero normalized values, highlighting significant mineralization patterns. Second, we overlay the mineralogical images on a grayscale geoglogical image to provide geographic context. Finally, we apply a rainbow colormap to the normalized data, where warmer colors indicate higher mineralization potential. This facilitates an intuitive interpretation of mineralization intensity and distribution.
The resultant visualization enables a clear understanding of potential copper deposits' intensity and spatial distribution through its color-coded representation.

\subsection{Data Validation}
\label{sec:validation}
%用人类识别的准确率以及结合最终的图像来说明是识别以及处理过程中的准确率
\begin{figure*}
    \centering
    \includegraphics[width=0.9\linewidth,trim=0pt 110pt 80pt 0pt, clip]{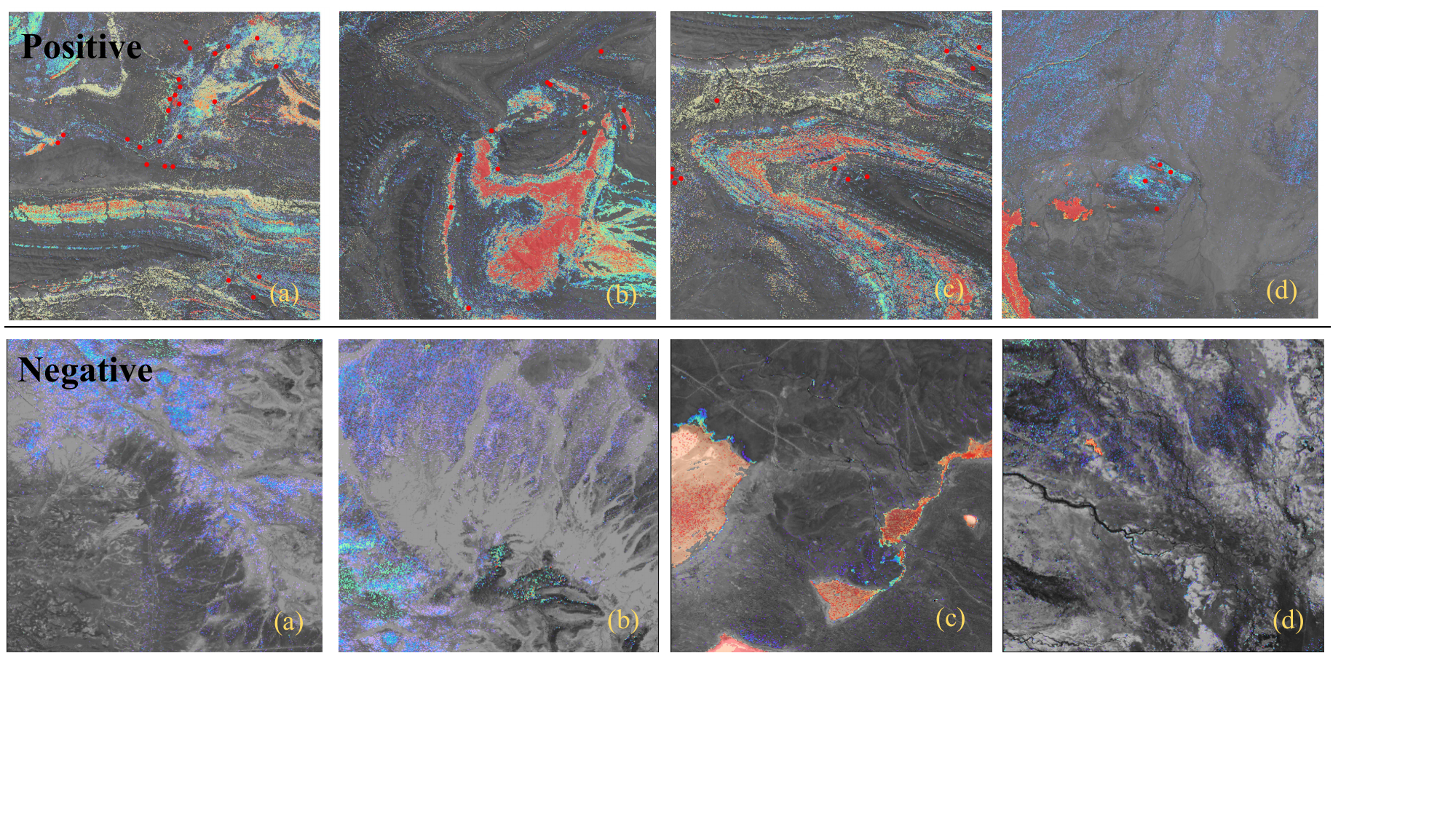}
    \caption{Data Validation. Comparing synthetic mineralization patterns with ground truth deposit locations.}
    \label{fig:verification}
\end{figure*}
By carefully mapping color transitions from red to yellow to orange and modulating spatial scales, we created a nuanced visual representation of potential mineral deposits.
To ensure the reliability of our synthetic data, we conducted a rigorous human-verified visual assessment. Expert manually examined the spatial and chromatic characteristics, comparing our synthetic mineralization patterns with ground truth deposit markers. This meticulous verification process confirmed the high correlation between our synthetic representations and actual mineral deposit locations.

As shown in  Figure~\ref{fig:verification}, the synthetic mineral prospectivity maps reveal distinct patterns of mineral potential. The authentic deposit locations (marked in red points) predominantly align with synthetic color-coded regions. 
Positive samples exhibit a gradual color transition from red to yellow to orange, systematically capturing the alteration zone characteristic of complete mineral deposits and reflecting the continuous mineralization processes. In contrast, negative samples display markedly different characteristics, with either low color intensity indicating minimal mineralization potential or an absence of complete color transitions suggesting incomplete deposit formation.

For example, the positive sample (d) demonstrates a subtle but critical color gradient transitioning, capturing the delicate mineralization patterns of mineral deposit. The scale of this sample, though small, precisely matches the actual deposit location markers.
In contrast, the negative samples (c) and (d) exhibit significant differences. These regions feature large-scale, high-intensity red areas that lack the nuanced color transitions and the unnatural spatial distribution, immediately signaling these as potentially unreliable mineral deposit indicators.

\section{Score Distribution of MLLMs and Human Assessment}
\label{sec:score_distribution}
\begin{figure*}
    \centering
    \includegraphics[width=0.9\linewidth]{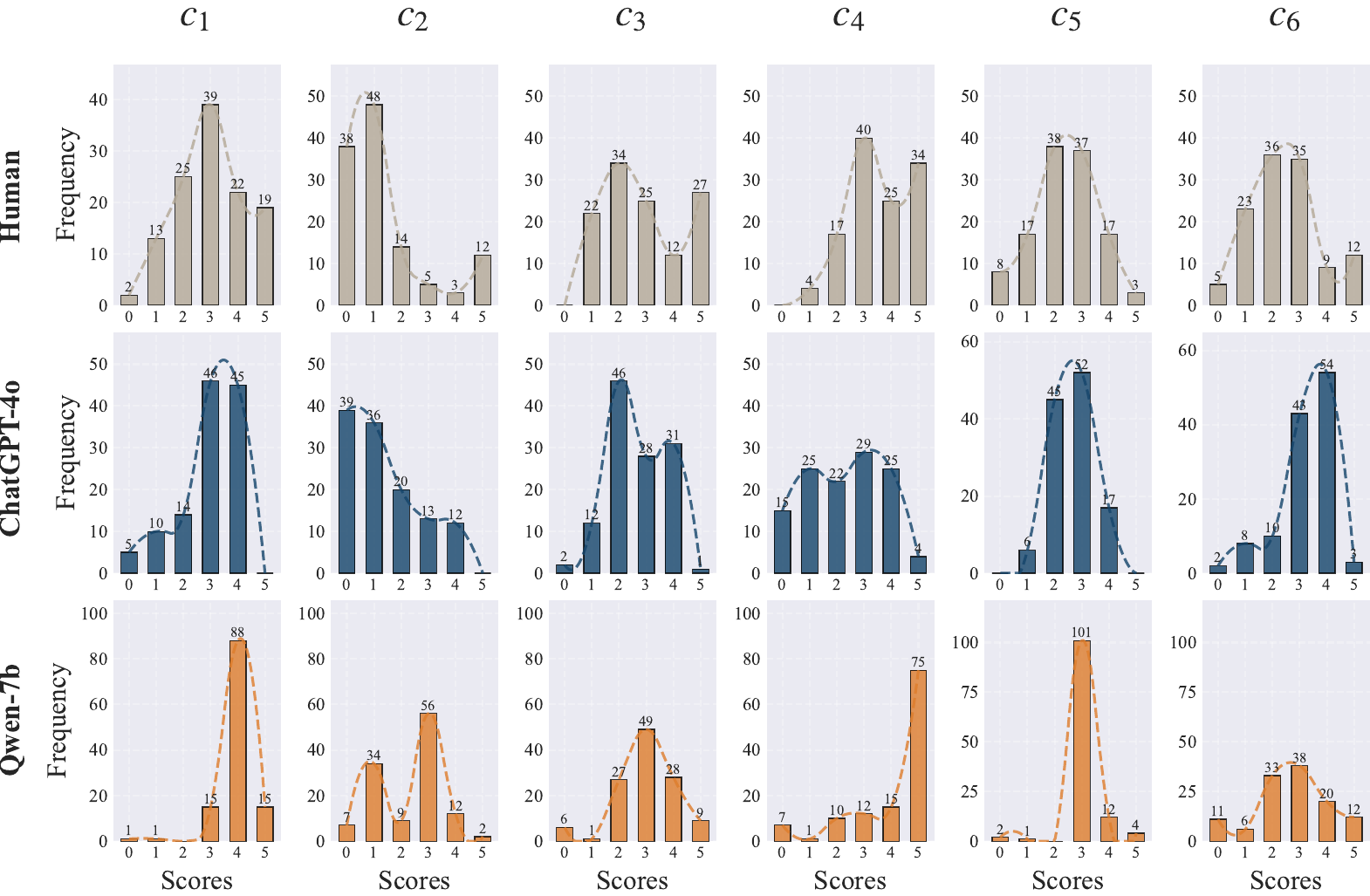}
    \caption{Score Distribution of MLLMs and Human Assessment}
    \label{fig:score_dis}
\end{figure*}
Based on the score distribution visualization in Figure \ref{fig:score_dis}, we observe distinct patterns across human assessments and the two models (GPT-4o and Qwen-7b). 
GPT-4o's score distribution closely aligns with human assessments, showing a more balanced and diverse distribution across different score levels. 
In contrast, Qwen-7b's scores tend to concentrate around a single value, as evidenced by the sharp peaks in its distribution. 
This indicates a notable limitation of Qwen-7b in capturing nuanced distinctions in evaluation criteria, highlighting its reduced variability and less human-like reasoning capability compared to GPT-4o.
\begin{figure*}[htbp]
\centering
\includegraphics[width=0.9\linewidth, trim=0pt 50pt 140pt 0pt, clip ]{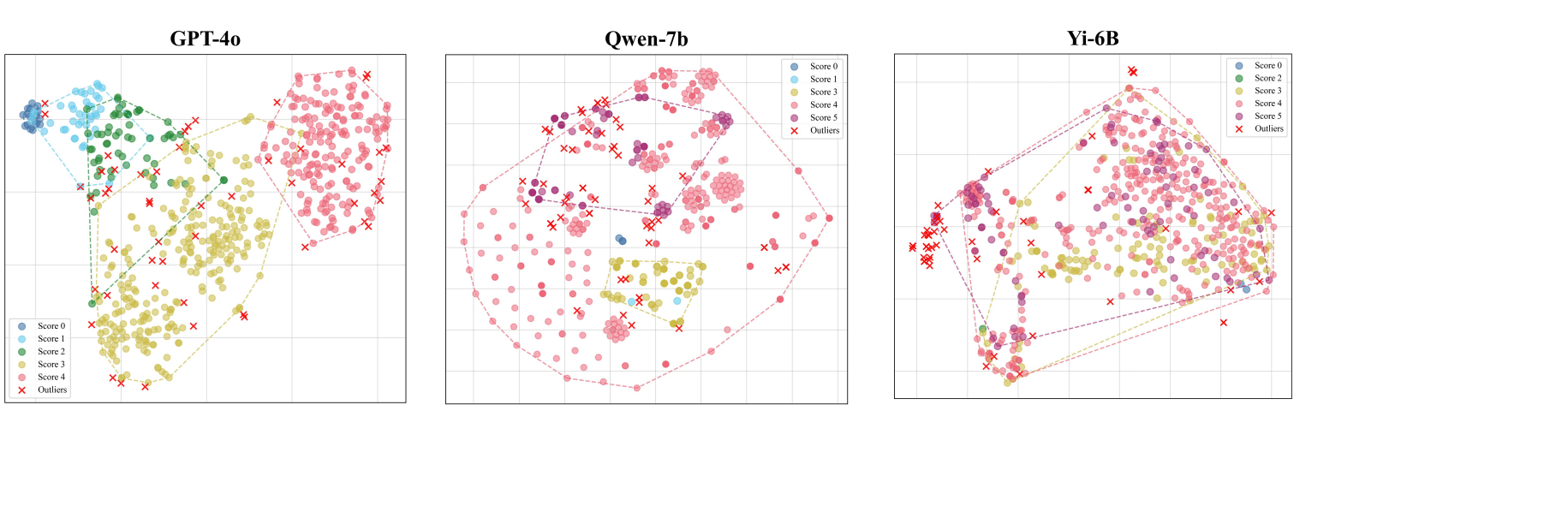}
\caption{Visualization of score-explanation alignment using T-SNE projection. Data points are color-coded by scores ranging from 0 to 5, with red crosses marking statistical outliers. Ideally, the explanation should be consistent with the assigned scores, leading to the clustering patterns.}
\label{fig:text-score}
\end{figure*}

\section{Impact of Decision-making Modules}
\label{sec:weight}
%只使用主要因素和全部因素之间的准确率对比，表格总结
\begin{table}[t]
\centering
\small
\renewcommand{\arraystretch}{1.4} % Adjusts row height
\setlength{\tabcolsep}{2.2pt} % Adjusts column width
\begin{tabular}{c c c c c c }
\hline
Setting & weights   & F1 (Pos.) & F1 (Avg.) & Roc-Auc & Mcc \\ \hline
\multirow{3}{*}{Hard}   & Local       & 24.86     & 34.4   & 58.81   & 12.94\\ 
                        & Mean   &31.58     & 56.33  & 64.81   & 20.65 \\ 
                        & Automatic     & 36.51     & 60.47   & 68.63   & 27.07 \\ 
                        \hline
\multirow{3}{*}{Standard} & Local       & 31.82     & 49.6   & 69.03   & 24.71 \\ 
                        & Mean   & 39.44     & 63.36  & 69.71   & 30.41 \\ 
                        & Automatic     & 61.20     & 77.19   & 83.35   & 56.30 \\ 
                        \hline
\multirow{3}{*}{Easy}   & Local   & 62.75     & 77.65  & 87.62   & 59.46 \\ 
                        &  Mean   & 67.13     & 81.40  & 80.84   & 62.82 \\ 
                        & Automatic    & 71.62     & 83.86   & 84.26   & 67.73 \\ 
                        \hline
\end{tabular}
\caption{Performance across different settings and weight allocation strategies.}
\label{tab:weights}
\end{table}
\paragraph{The Impact of Assessment Tuples.}  
Multiple criteria in the assessment tuples $ C = \{c_1, c_2, \ldots, c_M\} $ are used as input in the decision-making modules to guide the final decision $ o^{\textbf{(dm)}}$, as shown in Eq.\ref{equ:decisionmaking_modules}. The various criteria contribute to robust reasoning compared to using a single criterion. To evaluate their impact$ o^{\text{(dm)}} $, we employ three weighting strategies: 
(1) \emph{Local} which focuses on three data-driven types: geological context $c_1$, hyerspectral context $c_5$ and cross-reference validation $ c_6 $ without intermediate results $c_2$--$c_4$;  
(2) \emph{Mean} which assigns equal weights to $ c_1$--$ c_6 $; and  
(3) \emph{Automatic}, which dynamically balances criteria using Bayesian optimization.  

As shown in Table~\ref{tab:weights}, the \emph{Local} setting obtained the worst accuracy due to the exclusion of $c_2$--$c_4$. In contrast, the \emph{Mean} setting improved overall performance but failed to capture information hierarchies effectively. The \emph{Automatic} setting achieved superior results through the dynamic integration of all available outputs, effectively ensuring high-confidence answers.  
These findings demonstrate that the decision-making modules enhance model robustness by considering multiple assessment criteria in a balanced manner.

\paragraph{The Parameters of Decision-making Modules.}
\label{sec:expriment_setup}
%不同因素在不同任务对于模型的影响
\begin{figure*}[htbp]
    \centering
    \includegraphics[width=1\linewidth]{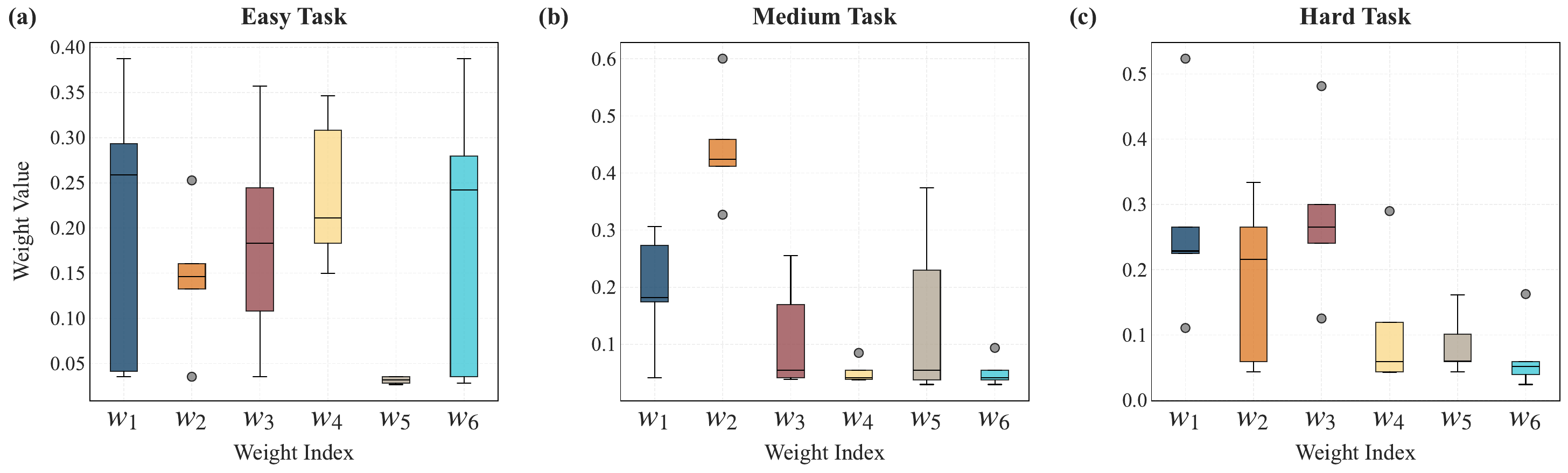}
    \caption{The Parameters of Decision-making Modules. $w_1$ to $w_6$ are assigned to various criterias ($c_1$ to $c_6$) across three task settings. Each box represents the range of weight values.}
    \label{fig:box}
\end{figure*}
We employed a five-fold cross-validation approach combined with Bayesian optimization to calculate the parameters weights $w_i$ for combining $\gC$. The average weights from the five validation folds were used as the final parameters for evaluating all MLLMs. 
The threshold used to generate a classification label is set 3 in all evaluation.
The final weight parameters and their distributions across the five folds are shown in Figure~\ref{fig:box}. This figure illustrates how the model adjusts its reliance on different criteria as task complexity increases. Each setting reveals unique patterns, highlighting the adaptive output of the decision-making module in response to varying levels of difficulty.
In ``Easy'' tasks, the weights $ w_1 $ to $ w_4 $ for criteria $ c_1 $ to $ c_4 $ are relatively uniform, ranging from approximately 0.15 to 0.3. This balanced distribution reflects the model's reliance on these single-category criteria for straightforward reasoning. The cross-image criterion $ c_5 $ is assigned negligible weight, indicating its minimal contribution. This uniformity suggests that higher-order information integration is unnecessary for simple tasks, where single-category criteria alone suffice for accurate inference.

As tasks increase to ``Standard'' complexity, the weight distribution shifts toward more discriminative features. For instance, $w_2$ increases significantly to approximately 0.5, underscoring its role as a key criterion. While cross-image criteria gain slightly more weight, they remain secondary as the model integrates aggregated insights alongside foundational criteria. Notably, single-category criteria continue to dominate the reasoning process in this setting.

In ``Hard'' tasks, the model heavily relies on a few single-category criteria (e.g., $ w_1 $ and $ w_3 $), while cross-image criteria ($ w_5 $ and $ w_6 $) maintain negligible contributions. This reflects their reduced utility as task complexity increases, further emphasizing the importance of individual, highly discriminative criteria for complex reasoning.

\section{Score-Explanation Consistency}
\label{sec:score_explanation_consistency}
To evaluate the MLLMs' reasoning ability, we analyzed the alignment between scoring decisions $s_i$ and explanations $e_i$ in Eq.~\ref{equ:output_format} across three representative MLLMs: GPT-4o, Qwen-7B, and Yi-6B. Using t-SNE~\cite{van2008visualizing} for visualization, the explanations are projected into BERT embeddings, with scores used as their labels for assessment. Based on identical scoring guidelines in the communication protocol, explanations $e_i$ should be aligned with the scores $s_i$, resulting in a compact clustering pattern. Conversely, scattered distributions may indicate potential inconsistencies in the judging process~\cite{du2024haloscope}.

As shown in Figure \ref{fig:text-score}, GPT-4o achieves superior clustering coherence, with distinct score-based clusters showing minimal boundary overlap. This clear structure indicates a strong alignment between scores and justifications, reflecting consistent reasoning patterns. Qwen-7B shows moderate clustering performance with distinguishable score groups but significant overlap in high-score regions (4-5), suggesting insufficient differentiation.
Yi-6B exhibits the most dispersed distribution with minimal explanation-group separation, indicating weak alignment between scores and explanations.
These clustering patterns correlate strongly with overall model performance, supporting our hypothesis that advanced models maintain more consistent score-explanation relationships.

\section{Effectiveness of Cross-image Tools}
\label{integration}

\begin{figure}[htbp]
\centering
\includegraphics[width=1\linewidth]{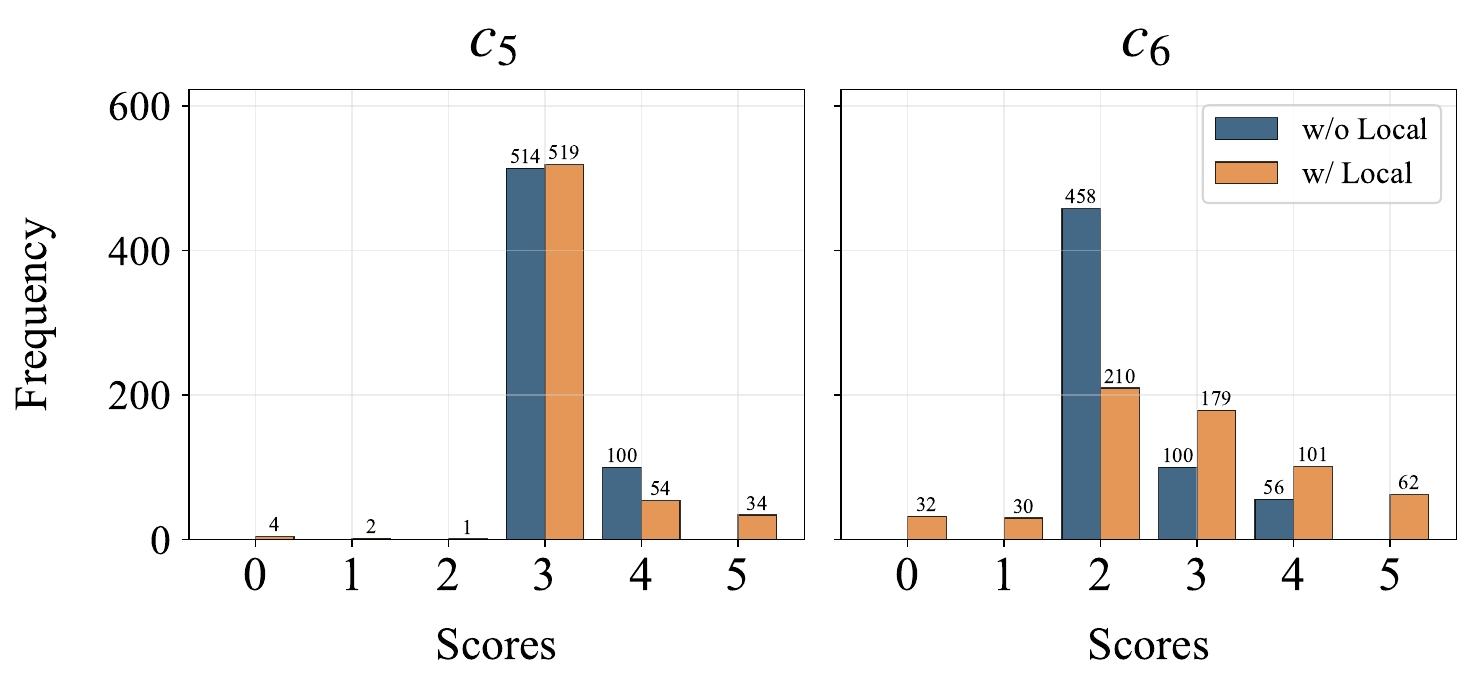}
\caption{Effectiveness of cross-image tools. The impact of including/excluding additional references as input on the score distribution for global criteria \( c_5 \) and \( c_6 \), evaluated using Qwen-7B.}
\label{fig:prior}
\end{figure}
As shown in Table~\ref{tab:tools_summary}, cross-image tools leverage inferences from single-category tools to analyze complex relationships between multiple images. This integration harnesses complementary strengths, resulting in a balanced score distribution that reflects improved model robustness across diverse inference tasks.
Figure~\ref{fig:prior} demonstrates that removing single-category inferences leads to a more concentrated score distribution, indicating that the model struggles to capture nuanced image features without auxiliary inputs. In contrast, integrating these inferences significantly improves the score distribution by increasing the proportion of high-scoring regions and reducing low-scoring instances. This improvement highlights enhanced reasoning accuracy.
These findings emphasize the critical role of hierarchy flow in judging modules in strengthening inference robustness. By combining both global and detailed perspectives, the judging modules establish a more reliable reasoning for addressing cross-image reasoning challenges.

% \section{Error Analysis}
% \label{error}

\section{The Exploration Pipeline and Case Study}
\label{sec:case-study}
\begin{figure*}[htbp]
    \centering
    \includegraphics[width=1.15\linewidth, trim=0pt 120pt 0pt 0pt, clip]{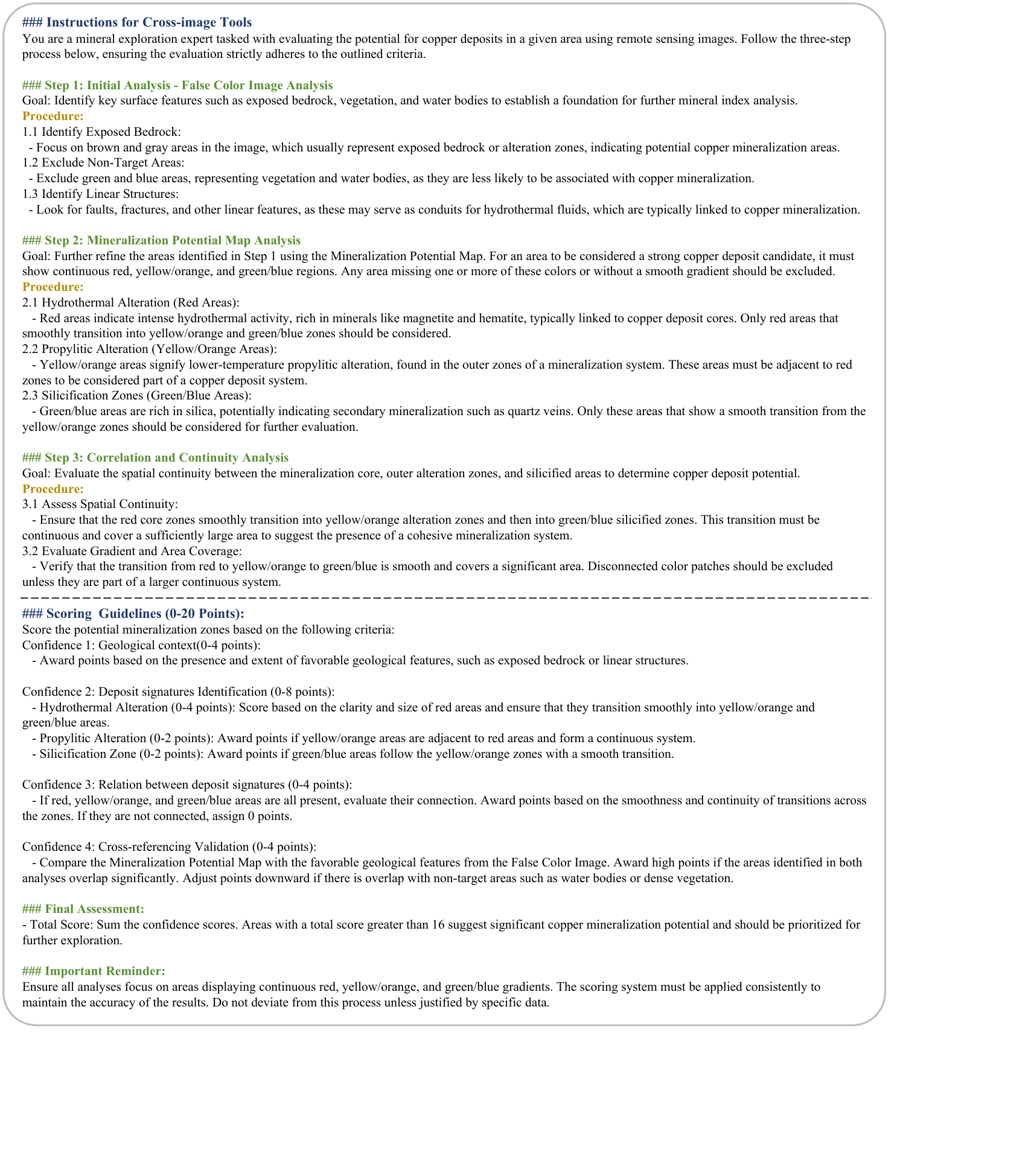}
    \caption{The pipeline of ``Easy'' setting}
    \label{fig:prompt-easy}
\end{figure*}

\begin{figure*}[htbp]
    \centering
    \includegraphics[width=1.15\linewidth, trim=0pt 0pt 0pt 0pt, clip]{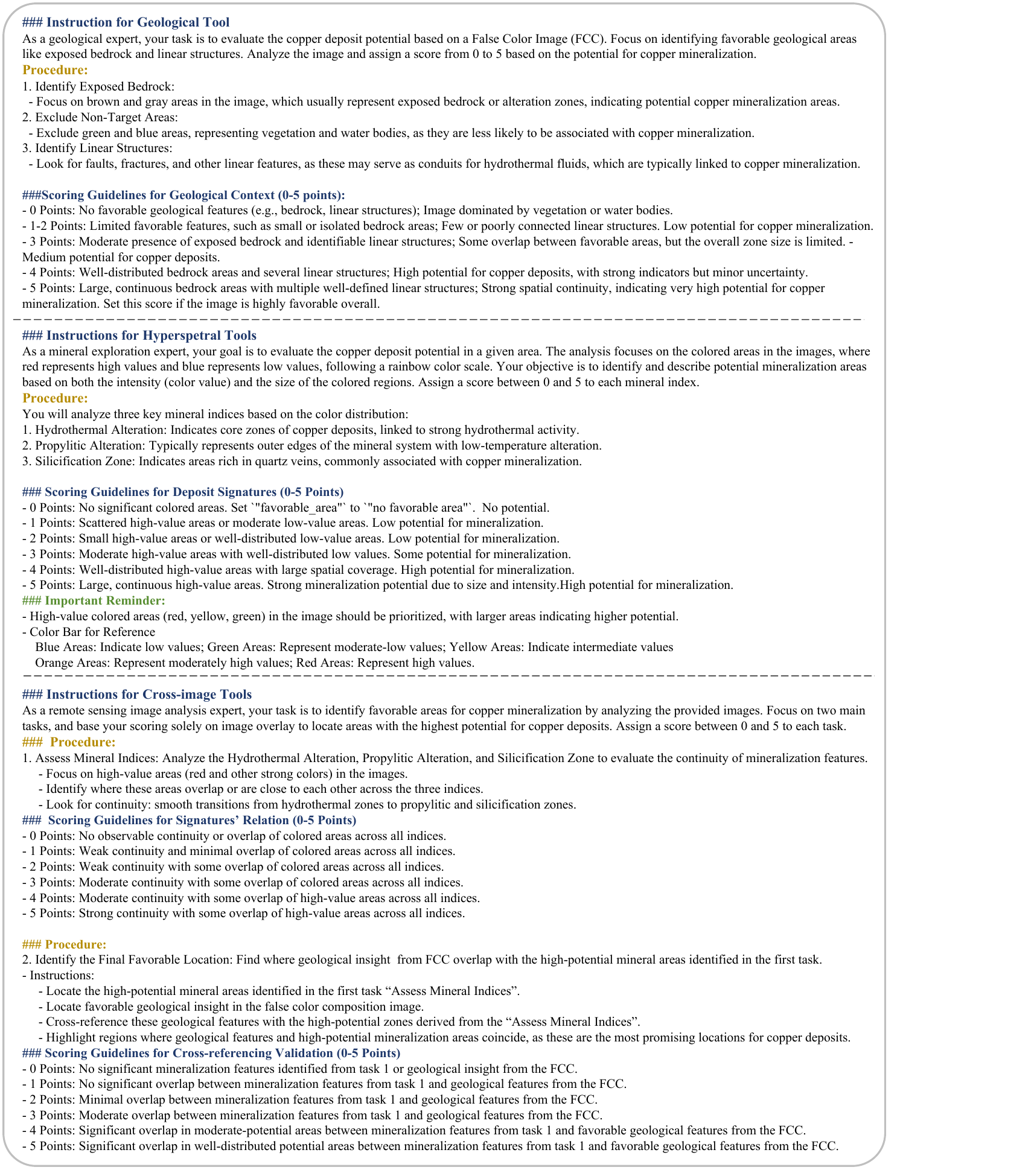}
    \caption{The pipeline of ``Standard'' setting}
    \label{fig:prompt-medium}
\end{figure*}

\begin{figure*}[tbp]
    \centering
    \includegraphics[width=1.15\linewidth, trim=0pt 285pt 0pt 0pt, clip]{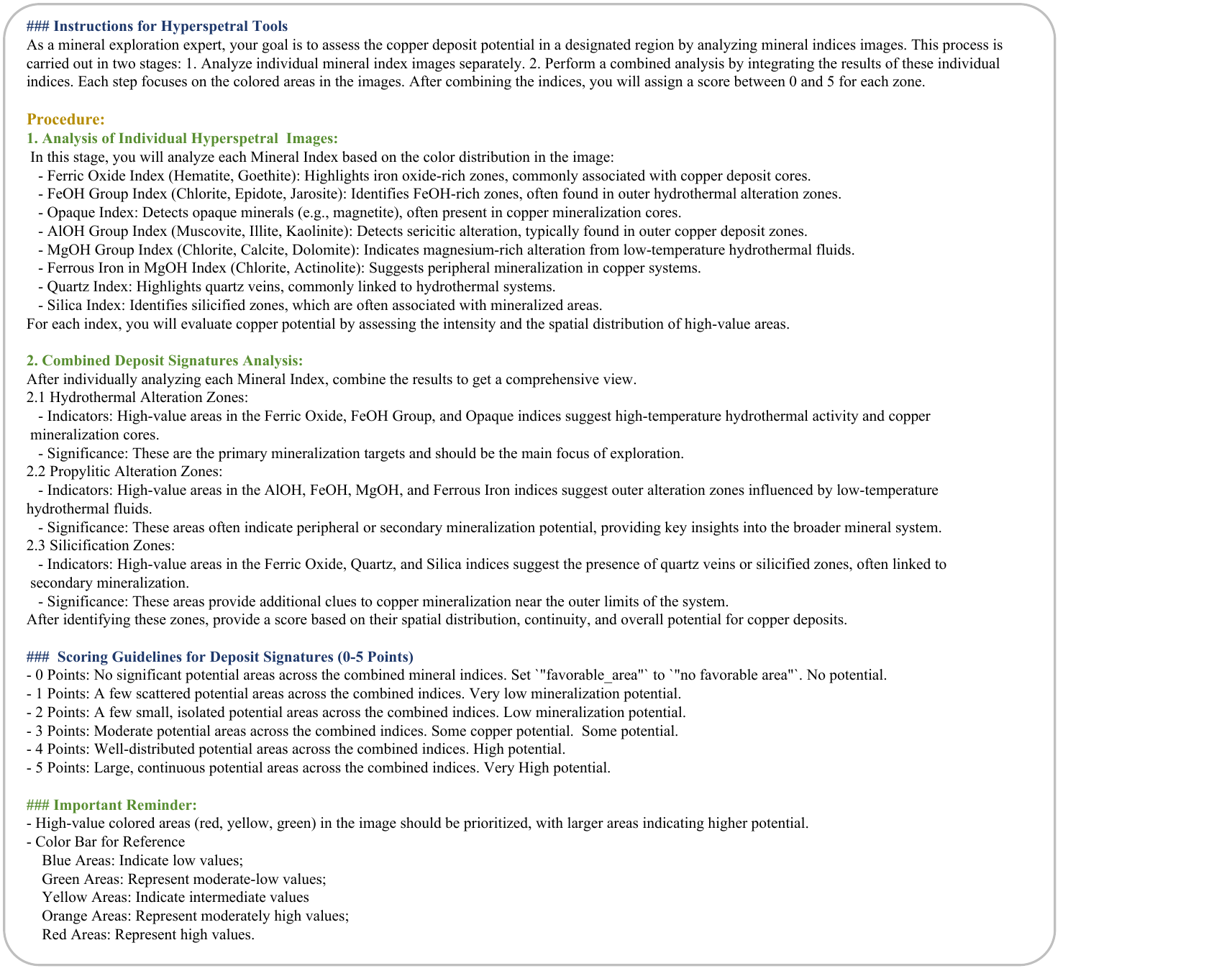}
    \caption{The pipeline of ``Hard'' setting}
    \label{fig:prompt-hard}
\end{figure*}

\begin{figure*}[htbp]
    \centering
    \includegraphics[width=0.95\linewidth, trim=0pt 800pt 0pt 0pt, clip]{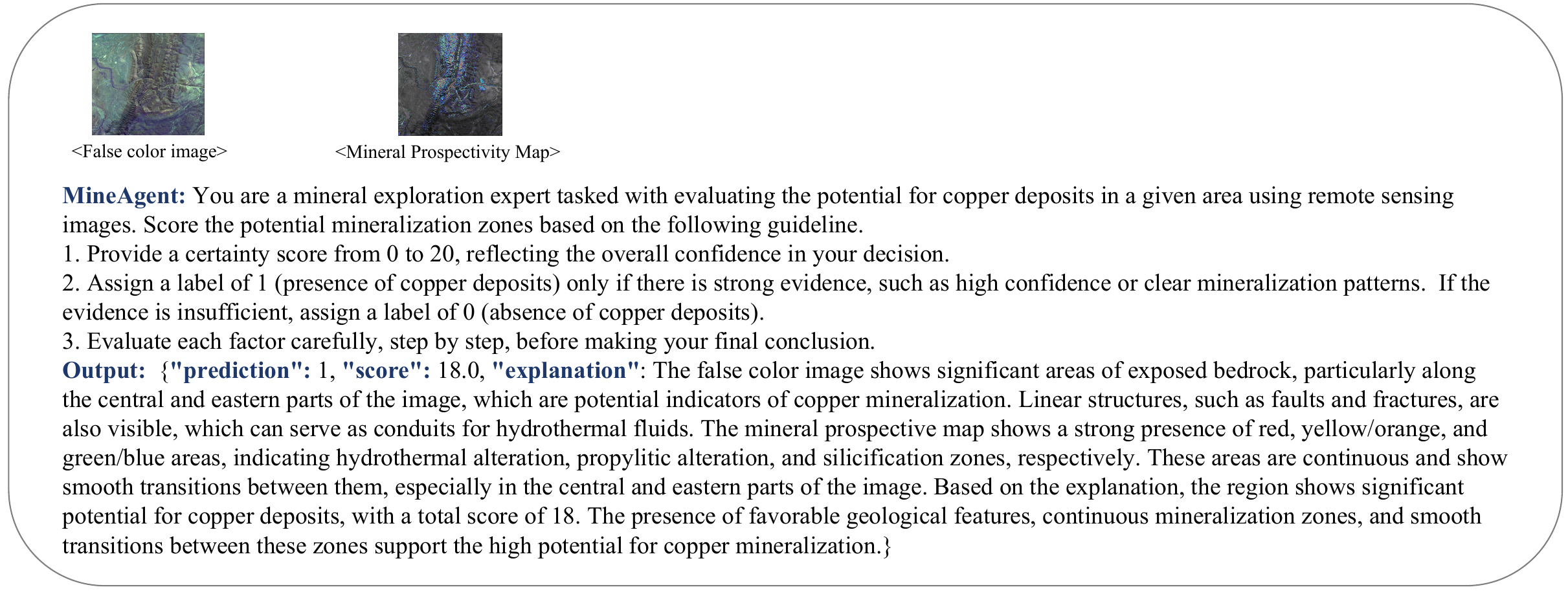}
    \caption{The example of ``Easy'' setting}
    \label{fig:example-easy}
\end{figure*}

\begin{figure*}[htbp]
    \centering
    \includegraphics[width=0.95\linewidth, trim=0pt 20pt 0pt 0pt, clip]{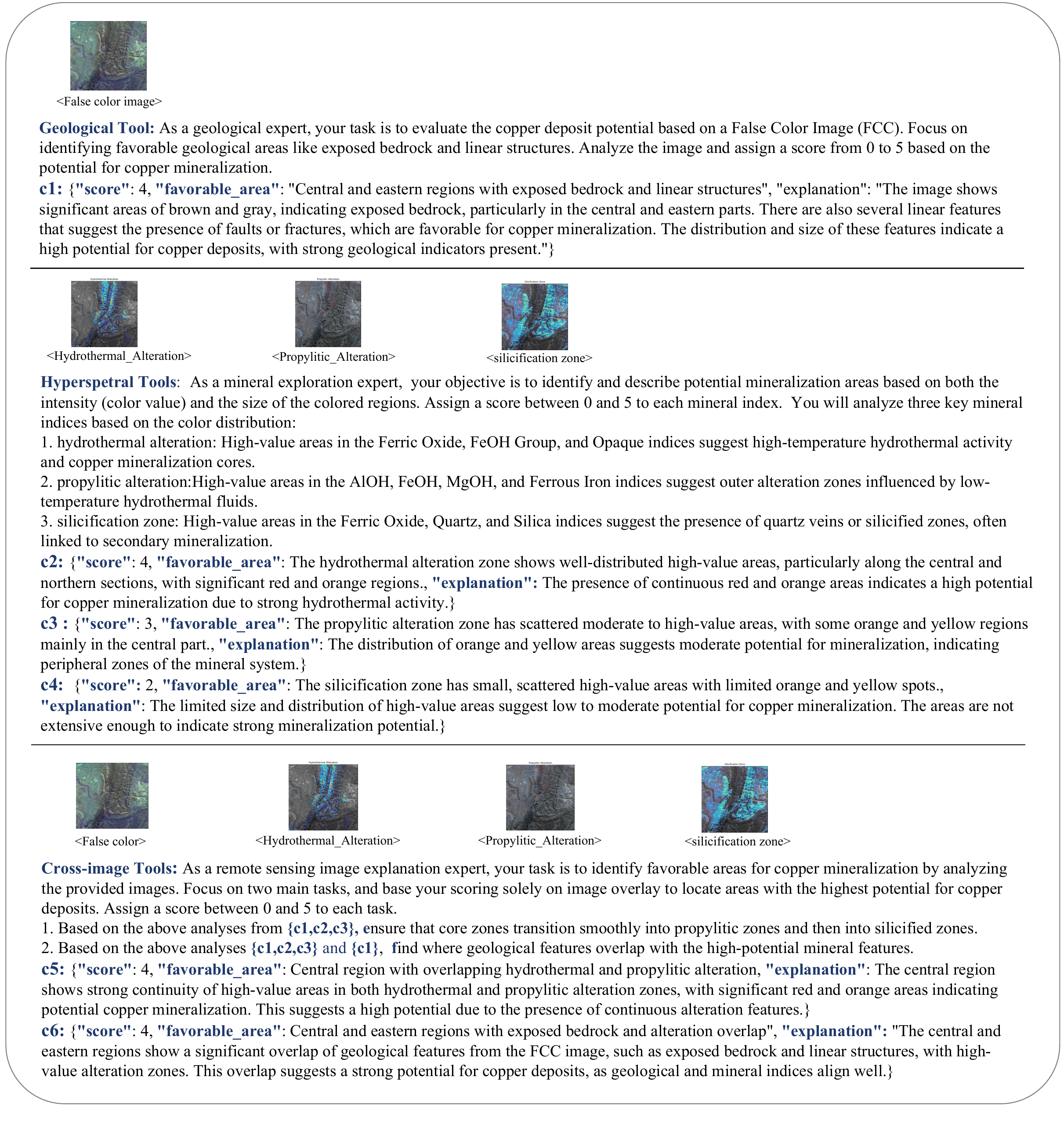}
    \caption{The example of ``Standard'' setting}
    \label{fig:example-medium}
\end{figure*}

\begin{figure*}[htbp]
    \centering
    \includegraphics[width=0.95\linewidth, trim=0pt 130pt 0pt 0pt, clip]{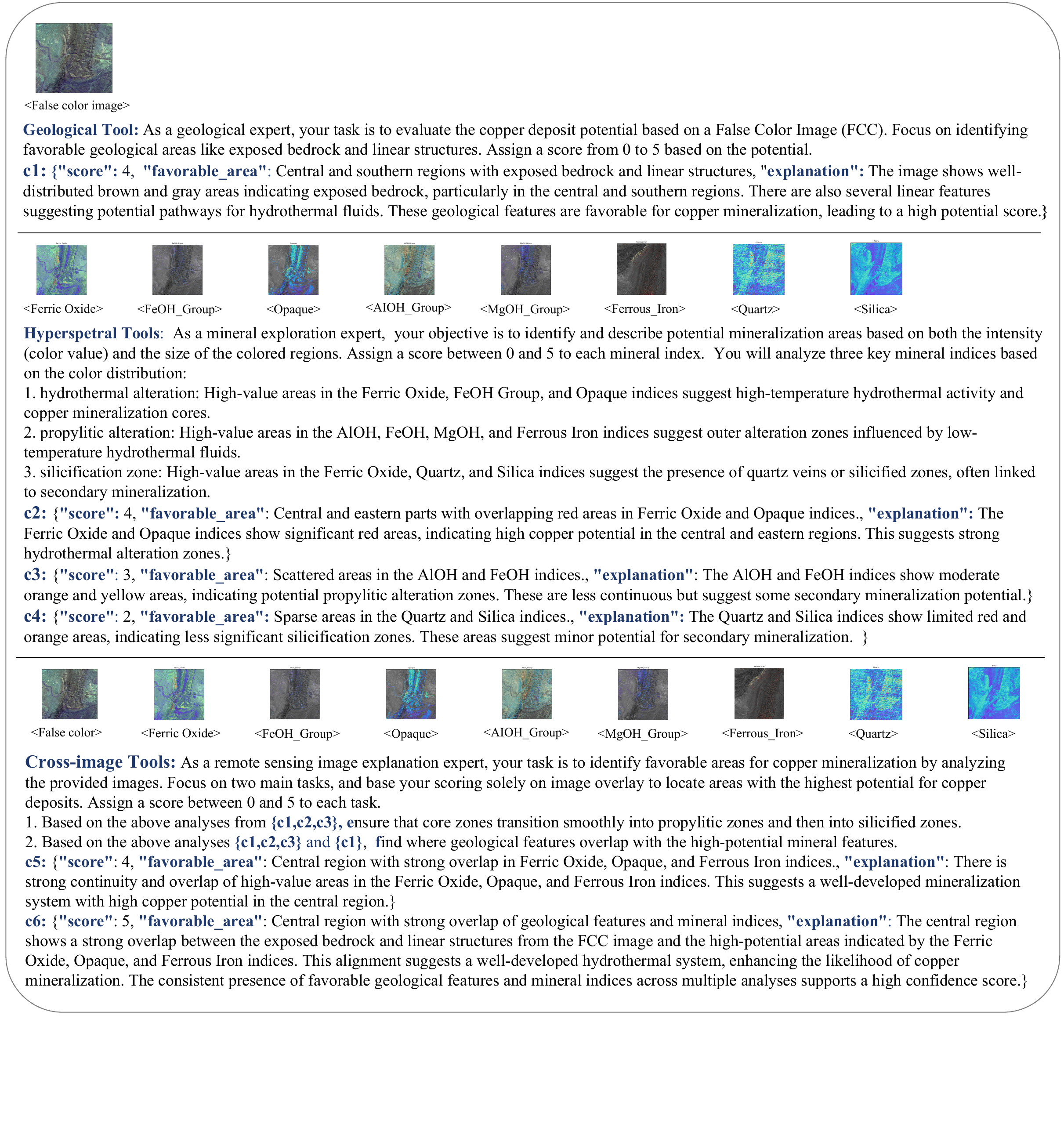}
    \caption{The example of ``Hard'' setting}
    \label{fig:example-hard}
\end{figure*}

% prompt
The detailed pipelines for the ``Easy,`` Standard,'' and ``Hard'' settings are shown in Figure~\ref{fig:prompt-easy} \ref{fig:prompt-medium} and \ref{fig:prompt-hard}, respectively.
The ``Standard'' and ``Hard'' pipelines differ only in the design of the hyperspetral tools. Consequently, the ``Hard'' pipeline focuses exclusively on the hyperspetral tools segment. For each setting, a case study is provided to illustrate the reasoning process in action, as shown in Figure  \ref{fig:example-easy} \ref{fig:example-medium} and \ref{fig:example-hard}.

% example

\section{More Related Work}
\paragraph{Machine Learning for Mineral Exploration.}
Mineral exploration represents a complex classification problem in geoscience, integrating diverse data sources to predict the location, quantity, and quality of mineral deposits~\cite{carranza2008geochemical}. 
Over the years, machine learning (ML) has played a transformative role in this field. Traditional approaches, such as principal component analysis~\cite{Ousmanou2024Mapping}, k-means~\cite{ren2020improved}, and regression trees~\cite{pham2018spatial}, have shown success in identifying prospective mineral deposits.
With the increasing complexity of remote-sensing data, deep learning (DL) models have demonstrated efficient and accurate feature extraction capabilities~\cite{liu2023deep,8}.
However, existing ML and DL methods face key limitations, including dependence on labeled datasets and poor generalizability across diverse geological environments.
To address these challenges, we propose leveraging advanced MLLMs to explore mineral deposits, integrating human expertise for enhanced insights.

\end{document}